\documentclass[lettersize,journal]{IEEEtran}
\usepackage{array}
\usepackage[caption=false,font=normalsize,labelfont=sf,textfont=sf]{subfig}
\usepackage{textcomp}
\usepackage{stfloats}
\usepackage{url}
\usepackage{verbatim}
\usepackage{amsmath,amsfonts}
\usepackage{tikz}
\usetikzlibrary{matrix}
\usepackage{import}
\usepackage{float}
\usepackage{mathtools}
\usepackage{cite}

\usepackage{graphicx}

\usepackage{bm}
\usepackage{hyperref}
\usepackage{newtxmath} 

\usepackage[noabbrev,nameinlink]{cleveref}

\usepackage[dvipsnames]{xcolor}
\definecolor{mycolor1}{RGB}{24, 92, 68}

\hypersetup{
    colorlinks = true,
    linkcolor = mycolor1,   
    citecolor = mycolor1,   
    urlcolor = Blue          
}

\usepackage{orcidlink} 

\newcolumntype{C}[1]{>{\centering\arraybackslash}m{#1}} 

\DeclareMathOperator*{\diag}{diag}
\DeclareMathOperator*{\card}{card}

\DeclareUnicodeCharacter{2217}{\textasteriskcentered}

\DeclareSymbolFont{AMSb}{U}{msb}{m}{n}
\DeclareSymbolFontAlphabet{\mathbb}{AMSb}
\begin{document}

\title{\LARGE \bf Optimal Control of Microswimmers for Trajectory Tracking Using Bayesian Optimization}

\author{Lucas Palazzolo$^{1,*}$ \orcidlink{0009-0002-4188-014X}, Mickaël Binois$^{2,\ddagger}$ \orcidlink{0000-0002-7225-1680} and  Laetitia Giraldi$^{1,\dagger}$\orcidlink{0000-0003-2684-0203}
\thanks{$^{1}$Université Côte d'Azur, Inria, Calisto team, Sophia Antipolis, France}%
\thanks{$^{2}$Université Côte d'Azur, Inria, Acumes team, CNRS, LJAD, Sophia Antipolis, France}%
\thanks{$^{*}$ {\tt\small lucas.palazzolo@inria.fr}}%
\thanks{$^{\dagger}$ {\tt\small laetitia.giraldi@inria.fr}}%
\thanks{$^{\ddagger}$ {\tt\small mickael.binois@inria.fr}}}


\maketitle

\begin{abstract}
Trajectory tracking for microswimmers remains a key challenge in microrobotics, where low-Reynolds-number dynamics make control design particularly complex. In this work, we formulate the trajectory tracking problem as an optimal control problem and solve it using a combination of B-spline parametrization with Bayesian optimization, allowing the treatment of high computational costs without requiring complex gradient computations.
Applied to a flagellated magnetic swimmer, the proposed method reproduces a variety of target trajectories, including biologically inspired paths observed in experimental studies. We further evaluate the approach on a three-sphere swimmer model, demonstrating that it can adapt to and partially compensate for wall-induced hydrodynamic effects. 
The proposed optimization strategy can be applied consistently across models of different fidelity, from low-dimensional ODE-based models to high-fidelity PDE-based simulations, showing its robustness and generality.
These results highlight the potential of Bayesian optimization as a versatile tool for optimal control strategies in microscale locomotion under complex fluid–structure interactions.
\end{abstract}


\section{Introduction}
\IEEEPARstart{M}{icroswimmers} are microorganisms using self-propulsion or external actuation in fluid environments, including sperm cells, bacteria, and bio-inspired microrobots. There is considerable interest in understanding and controlling such swimmers, particularly for medical applications such as targeted drug delivery \cite{medical_robots_cancer_2020, palagi2018}.\\

These systems operate in the Stokes regime due to their small size, corresponding to low Reynolds numbers. In this regime, viscous forces dominate inertial forces, so that reciprocal motions do not generate net displacement—a phenomenon known as Purcell's \textit{scallop theorem}. Flagellated swimmers, composed of a head and an elastic flagellum, achieve propulsion through time-irreversible deformations, as exemplified by spermatozoa \cite{jikeli2015} and other flagellated microorganisms. Rigid swimmers, such as the three-sphere model introduced by Najafi and Golestanian \cite{najafi2004}, rely on sequences of non-reciprocal actuations to break time-reversal symmetry of Stokes flows. The controllability and optimal actuation of such microswimmers have been extensively studied, including for flagellar deformations \cite{marchello2022, giraldi2013, zoppello2025}, magnetically actuated swimmers \cite{moreau2019}, and in the presence of boundaries (see \cite{alouges2013} and \cite{gerardvaret2015}). The search of control (optimal or not) has been extensively studied, both experimentally \cite{grosjean2015,cheang2017}, and numerically using various optimization methods from classical deterministic methods \cite{giraldi2013, faris2020} to machine learning-based techniques \cite{elKhiyati2023}. \\

Recently, Bayesian optimization (BO) has emerged as a powerful technique for solving high-cost or black-box optimization problems, particularly in design and shape optimization contexts where evaluations are expensive and gradients are unavailable \cite{Frazier2018,garnett2023}. 
In recent years, BO has been applied to various applications, including aerospace engineering \cite{Lam2018}, hyperparameter tuning in machine learning models \cite{snoek2012} and in robotics for multiple applications (\cite{Rai2019} for bipedal robots, \cite{Nogueira2016} for grasping robot, \cite{Zheng2023} for coordinating multi-robots). 
Despite these advances in robotics, the application of BO to microscale swimming control remains unexplored. Recent studies have started to employ surrogate-based optimization techniques for the design of low–Reynolds-number systems \cite{palazzolo2025a}. However, the use of BO in a control-oriented setting for microswimmer locomotion has not yet been systematically studied. To our knowledge, BO has not previously been applied to microswimmer control. In this work, we introduce the first application of BO to the control of microswimmers, connecting modern machine learning methods with the constraints of low–Reynolds-number locomotion.\\

More precisely, we study the use of BO to design control strategies for trajectory tracking of microswimmers. The highly nonlinear and computationally expensive cost of the underlying dynamics makes BO particularly suitable. We formulate the control problem as an optimal control. To overcome the curse of dimensionality in high-dimensional spaces with multiple constraints, we employ the \textit{Scalable Constrained Bayesian Optimization} (SCBO) method \cite{eriksson_scalable_2021}, implemented in \texttt{Python} via the \texttt{BoTorch} library \cite{balandat2020botorch}. The controls are represented using B-spline curves, a flexible class of functions widely used in engineering and computer graphics, parameterized by a finite set of control points \cite{piegl1997}. Their smoothness, local support properties and finite-dimensional representation make them ideal for this kind of problem.\\

We consider two distinct microswimmer models. The first model is the $N$-links swimmer, which consists of a magnetic head and an elastic flagellum discretized into $N$ rigid links \cite{giraldi2013, alouges2015}. The forces acting on the links include hydrodynamic drag approximated by Resistive Force Theory (RFT) \cite{gray1955, friedrich2010, moreau2018}, elastic restoring forces, and magnetic torques on the head. This leads to an ordinary differential equation (ODE) model, which is solved in \texttt{Matlab}. The purpose is to make the design of closed-loop control strategies for trajectory tracking  \cite{oulmas2017}. The second model is the three-sphere swimmer, governed by the unsteady Stokes equations coupled with Newton's laws for rigid bodies. The resulting partial differential equation (PDE) system is solved in weak form using the finite element library \texttt{Feel++} (see \cite{berti2021} and \cite{prudhomme2025}).
The optimization of the three-sphere model has been extensively investigated in the literature. Various studies have focused on optimizing the stroke patterns, swimming efficiency, and energetics of low-Reynolds-number swimmers using analytical, numerical, and data-driven approaches \cite{alouges2008, Tam2007, Avron2004}. However, most of these investigations assume an unbounded fluid domain, neglecting the presence of boundaries that are often unavoidable in realistic microfluidic environments. While several studies have examined the impact of walls on microswimmer dynamics \cite{alouges2013, gerardvaret2015, spagnolie2012}, the role of boundaries in optimal control problems is not fully understood. Here, the optimization of the three-sphere swimmer's control in bounded domains is considered, where wall effects play a crucial role in propulsion and maneuverability. Finally, at the lowest fidelity level, we consider an ODE-based model in which fluid–structure interactions are fully approximated. At the other extreme, a detailed PDE-based model is employed, in which all physical forces are explicitly represented, including interactions with boundaries.\\

The resulting trajectories for the first model closely match the paths of a real sperm cell, as described in \cite{friedrich2010}, exhibiting periodic patterns.
For the second model, optimal trajectories that explicitly account for hydrodynamic wall effects were obtained. As expected, the swimmer’s motion forms double-loop trajectories near the boundary, indicating that compensating for wall-induced hydrodynamic effects requires a more complex gait. Within the geometric control framework, this behavior can be interpreted as the swimmer exploiting combinations of Lie brackets to achieve sufficient maneuverability within its reachable set.\\

The paper is organized as follows.  \Cref{sec:geo_nlinks} presents the general mathematical formulation of microswimming optimization as a constrained optimization problem with an abstract algebraic constraint. The two models studied in this work are introduced: the elastic flagellated microswimmer with a magnetic head and the three-sphere swimmer.  In \Cref{sec:ocp}, we describe the optimal control problems associated with each swimmer, as well as their numerical resolution using B-splines.  \Cref{sec:numres} provides numerical results for trajectory tracking with the $N$-link swimmer for different reference paths. Numerical results for the three-sphere swimmer are also presented, in particular regarding wall-effect compensation.  Finally, \Cref{sec:conclusion} summarizes and concludes the paper.

\section{Mathematical Modeling}\label{sec:mathmod}

In general, the dynamics of a microswimmer can be formulated as an abstract algebraic constraint. Let 
\begin{equation*}
(d_p,d_o,d_{\text{in}}) \in \{2,3\}\times \{1,2,3\}\times \mathbb{N},
\end{equation*}
and define the swimmer’s state by 
\begin{equation}\label{eq:swimmerstate}
\boldsymbol{p}(t) \in \mathbb{R}^{d_p} \times [0,2\pi]^{d_o} \times \mathbb{R}^{d_{\text{in}}},
\end{equation}
defined over a finite time horizon $T>0$ such as $t\in[0,T]$. Here, $d_p$ denotes the dimension of the swimmer’s position in physical space, $d_o$ the number of orientation variables in the laboratory reference frame, and $d_{\text{in}}$ a set of additional internal coordinates describing the swimmer’s configuration. The swimmer is actuated by a control vector
$\boldsymbol{u} \in L^{\infty}([0,T];\;\mathbb{R}^m)$ with $m\in\mathbb{N}^*$. For self-propelled swimmers, $\boldsymbol{u}$ represents intrinsic deformations such as flagellar bending (e.g., the $N$-link model \cite{giraldi2013}, Purcell’s three-link swimmer \cite{purcell}) or cyclic arm-length variations in the three-sphere swimmer of Najafi and Golestanian \cite{najafi2004}. In contrast, externally actuated swimmers are controlled by $\boldsymbol{u}$ through external stimuli such as magnetic fields \cite{alouges2015,Dreyfus2005,oulmas2017}, light \cite{varun2022}, or chemical gradients \cite{zhuang2017}.\\

\noindent To formalize the dynamics, let $\mathscr{P}$ denote the state space. For ODE-based models, one typically has $\mathscr{P} = \mathbb{R}^{d_p + d_o + d_{\text{in}}}$; for PDE-based models, $\mathscr{P}$ can be either $\mathbb{R}^{d_p + d_o + d_{\text{in}}}$ or a suitable function space such as $L^2(\Omega; \mathbb{R}^{d_p + d_o + d_{\text{in}}})$ or $H^1(\Omega; \mathbb{R}^{d_p + d_o + d_{\text{in}}})$, where $\Omega \subset \mathbb{R}^{d_p}$ denotes the fluid domain. Let $\mathscr{U} = \mathbb{R}^m$ denote the control space, and let $\mathscr{Y}$ be the residual space: for ODEs, $\mathscr{Y} = \mathbb{R}^n$, while for PDEs, $\mathscr{Y}$ is typically a function space defined over $\Omega$. Let $\mathscr{Z}$ denote the residual space associated with initial or boundary conditions.  We introduce two operators:
\begin{align*}
\boldsymbol{G} &: \mathscr{P} \times \mathscr{U} \times [0, T] \to \mathscr{Y}, && \text{(governing dynamics)},\\
\boldsymbol{C} &: \mathscr{P} \times [0, T] \to \mathscr{Z}, && \text{(initial and boundary constraints)}.
\end{align*}
The system is therefore expressed in the compact form:
\begin{align*}
\boldsymbol{G}(\boldsymbol{p}(t), \boldsymbol{u}(t), t) &= 0_{\mathscr{Y}}, \quad t \in [0,T],\\
\boldsymbol{C}(\boldsymbol{p}(t), t) &= 0_{\mathscr{Z}}, \quad t \in [0,T].
\end{align*}
The first equation enforces the swimmer’s and fluid’s dynamics, 
while the second enforces physical constraints such as initial and boundary conditions. 
For a given control $\boldsymbol{u}$, we denote by $\boldsymbol{p}_{\boldsymbol{u}}$ the corresponding state trajectory that satisfies these equations.\\

\noindent Within this abstract framework, problems of microswimmer design \cite{palazzolo2025b, ishimoto2016, shum2019} and control \cite{elKhiyati2023, faris2020, chambrion2019} can be written as constrained optimization problems. Given an admissible control set $\mathscr{U}_{\mathrm{ad}}$, one seeks to minimize a cost functional $J$:
\begin{equation}\label{eq:pb_opti_generic}
\begin{aligned}
    \inf_{\boldsymbol{u} \in \mathscr{U}_{\mathrm{ad}}} \quad &J(\boldsymbol{p}_{\boldsymbol{u}}, \boldsymbol{u})\\[2mm]
    \text{subject to} \quad 
    &\boldsymbol{G}(\boldsymbol{p}_{\boldsymbol{u}}(t), \boldsymbol{u}(t), t) = 0_{\mathscr{Y}},\\
    &\boldsymbol{C}(\boldsymbol{p}_{\boldsymbol{u}}(t), t) = 0_{\mathscr{Z}}, \quad \forall t \in [0,T].
\end{aligned}
\end{equation}

\noindent Here, the trajectory tracking task is formulated as an optimal control problem. Let the reference trajectory be denoted by $
\boldsymbol{p}_{\text{ref}} : [0,T] \to \mathbb{R}^3,
$ with final target state $\boldsymbol{p}_{\text{final}} = \boldsymbol{p}_{\text{ref}}(T)$. For a given control $\boldsymbol{u}$, we denote by $\boldsymbol{p}_{\boldsymbol{u}}$ the resulting trajectory, obtained either from the ODE model \eqref{eq:mat_form} in the case of the $N$-link swimmer, or from the PDE model \eqref{eq:three_sphere_pde} in the case of the three-sphere swimmer. The cost functional used to evaluate the tracking performance is defined as 
\begin{equation}\label{eq:cost_func}
J(\boldsymbol{p}_{\boldsymbol{u}},\boldsymbol{u})
    := \int_{0}^T \left\|\boldsymbol{p}_{\boldsymbol{u}}(t) - \boldsymbol{p}_{\text{ref}}(t)\right\|_{Q}^2 \, dt
    + \left\|\boldsymbol{p}_{\boldsymbol{u}}(T) - \boldsymbol{p}_{\text{final}}\right\|_{S}^2,
\end{equation}  
where $Q$ and $S$ are positive definite symmetric weighting matrices depending on the problem setting. The associated matrix norm is given by  
$\|\boldsymbol{p}\|_{Q} = \boldsymbol{p}^\top Q \boldsymbol{p}.$ The first term in \eqref{eq:cost_func} penalizes deviations from the reference trajectory along the time horizon, while the second term enforces accuracy at the final target state. The admissible control set $\mathscr{U}_{\text{ad}}$ is defined in \eqref{eq:adm_control_nlinks} and \eqref{eq:adm_control_threesphere}.

\subsection{$N$-links model}

\subsubsection{Geometrical modeling}\label{sec:geo_nlinks}

\begin{figure}[htpb]
    \centering
    \includegraphics[width=1\linewidth]{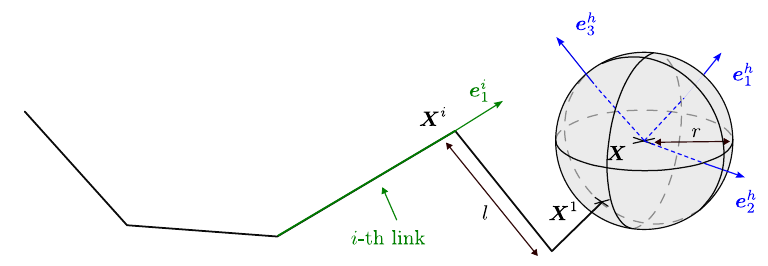}
\caption{3D $N$-links model. The swimmer's head frame is defined as $\mathscr{R}^{h} = (\boldsymbol{X}, \boldsymbol{e}_1^h, \boldsymbol{e}_2^h, \boldsymbol{e}_3^h)$. Each link $i$ of length $l$ is oriented along the unit vector $\boldsymbol{e}_1^i$ from $\boldsymbol{X}^i$. Taken from \cite{palazzolo2025b}}
    \label{fig:nlinks_illu}
\end{figure}

The swimmer, represented in \Cref{fig:nlinks_illu}, is modeled as a spherical head of radius $r$, centered at $\boldsymbol{X}\in\mathbb{R}^3$, to which a flagellum of length $L$ is attached. The flagellum is discretized into $N+1$ points $\boldsymbol{X}^i$ for $i \in \{1, \ldots, N+1\}$, connected by $N$ rigid links of length $l=L/N$. We denote by $\mathscr{R} = (0_{\mathbb{R}^3}, \boldsymbol{e}_1, \boldsymbol{e}_2, \boldsymbol{e}_3)$ the laboratory reference frame, and by $\mathscr{R}^{h} = (\boldsymbol{X}, \boldsymbol{e}_1^h, \boldsymbol{e}_2^h, \boldsymbol{e}_3^h)$ the reference frame attached to the swimmer’s head, where $(\boldsymbol{e}_1^h, \boldsymbol{e}_2^h, \boldsymbol{e}_3^h)$ forms an orthonormal basis.  The orientation of the swimmer is described by the rotation matrix $R^h \in SO(3)$, mapping vectors from $\mathscr{R}^{h}$ to the laboratory frame $\mathscr{R}$. We parameterize $R^h$ using three Tait–Bryan angles $(\theta_x, \theta_y, \theta_z)$ with the $ZYX$ convention, i.e., a rotation about the $z$-axis, followed by the $y$-axis, and finally the $x$-axis:  
\begin{equation}\label{eq:Rh}
    R^h := R_x(\theta_x) R_y(\theta_y) R_z(\theta_z),
\end{equation}
where $R_x$, $R_y$, and $R_z$ are the standard elementary rotation matrices about the $x$-, $y$-, and $z$-axes, respectively:
\begin{align*}
    R_x(\theta_x) &= \begin{bmatrix}
        1 & 0 & 0 \\
        0 & \cos\theta_x & -\sin\theta_x \\
        0 & \sin\theta_x & \cos\theta_x
    \end{bmatrix}, \\
    R_y(\theta_y) &= \begin{bmatrix}
        \cos\theta_y & 0 & \sin\theta_y \\
        0 & 1 & 0 \\
        -\sin\theta_y & 0 & \cos\theta_y
    \end{bmatrix}, \\
    R_z(\theta_z) &= \begin{bmatrix}
        \cos\theta_z & -\sin\theta_z & 0 \\
        \sin\theta_z & \cos\theta_z & 0 \\
        0 & 0 & 1
    \end{bmatrix}.
\end{align*}
For $i \in \{1, \ldots, N\}$, the direction of the $i$-th link, $\boldsymbol{e}_1^i$, is defined by the segment $[\boldsymbol{X}^{i+1}, \boldsymbol{X}^i]$, where the points $\boldsymbol{X}^i$ are given by
\begin{equation}\label{eq:Oi}
\begin{cases}
    \boldsymbol{X}^1 &= \boldsymbol{X} - r \boldsymbol{e}_1^h,\\
    \boldsymbol{X}^i &= \boldsymbol{X}^1 - l \displaystyle\sum_{k=1}^{i-1} \boldsymbol{e}_1^k, \quad \forall i \in \{2, \ldots, N+1\},
\end{cases}
\end{equation}
where $\boldsymbol{X}^1$ is the point at which the flagellum attaches to the head. To parameterize the vectors $\boldsymbol{e}_1^i$ in the head reference frame $\mathscr{R}^{h}$, we use spherical coordinates $(\phi_y^i, \phi_z^i) \in [0, 2\pi]^2$, corresponding to rotations about the axes $\boldsymbol{e}_2^i$ and $\boldsymbol{e}_3^i$, respectively. This choice is motivated by the assumption that the links have infinitesimal cross-sections (due to the RFT), so the rotation about $\boldsymbol{e}_1^i$ is negligible. Let $R^i$ denote the rotation matrix associated with these angles:
\begin{equation}\label{eq:Ri}
    R^i := R_y(\phi_y^i) R_z(\phi_z^i).
\end{equation}
The coordinates of the vector $\boldsymbol{e}_1^i$ in the laboratory frame $\mathscr{R}$ are then obtained by composing the head rotation $R^h$ and the link rotation $R^i$:
\begin{equation}\label{eq:ei}
    \boldsymbol{e}_1^i = R^h R^i \boldsymbol{e}_1.
\end{equation}
Thus, letting $s$ denote the arc length coordinate on the $i$-th link ($0\leq s \leq l$) and $\boldsymbol{x}^i(s)\in \mathbb{R}^{3}$ the coordinates associated with this arc length coordinate, we obtain for all $i \in \{1, \ldots, N\}$ and for all $s \in [0, l]$,
\begin{equation}
    \boldsymbol{x}^i(s) = \boldsymbol{X}^i - s R^h R^i \boldsymbol{e}_1.
\end{equation}
Using the properties \eqref{eq:idso3} and \eqref{eq:crossmatrix} of $SO(3)$, we can write $\dot{\boldsymbol{x}}^{i}(s)$ in the form
\begin{equation}\label{eq:dotxi}
    \dot{\boldsymbol{x}}^{i}(s) = \dot{\boldsymbol{X}}^i + s [R^hR^i \boldsymbol{e}_1]^{\times}\boldsymbol{\Omega}^h + s R^h[R^i\boldsymbol{e}_1]^{\times}\boldsymbol{\Omega}^i.
\end{equation}

\subsubsection{Swimmer dynamics}
The swimmer operates in a low Reynolds number regime, so inertia can be neglected. According to Newton's laws, the total force and torque acting on the swimmer must vanish. Let $\boldsymbol{F^{\text{hydro}}_{\text{head}}}$ (resp. $\boldsymbol{F}^{\text{hydro}}_i$) denote the viscous force exerted by the fluid on the head (resp. the $i$-th link), and $\boldsymbol{T^{\text{hydro}}_{\text{head}}}$ (resp. $\boldsymbol{T}^{\text{hydro}}_{i,\boldsymbol{x}^0}$) the viscous torque exerted on the head (resp. on the $i$-th link about $\boldsymbol{x}^0$). Let $\boldsymbol{F^{\text{ext}}}$ be the external force and $\boldsymbol{T_{\text{total}}^{\text{ext}}}$ (resp. $\boldsymbol{T}_i^{\text{ext}}$) the external torques applied to the head (resp. to link $i$). The dynamics of the swimmer are therefore governed by the system:
\begin{equation}\label{eq:newtonsyst}
    \begin{cases}
        \boldsymbol{F^{\text{hydro}}_{\text{head}}} + \displaystyle{\sum_{i=1}^N \boldsymbol{F}^{\text{hydro}}_i} &= \boldsymbol{F^{\text{ext}}}, \\
        \boldsymbol{T^{\text{hydro}}_{\text{head}}} + \displaystyle{\sum_{i=1}^N }\boldsymbol{T} _{i,\boldsymbol{X}}^{\text{hydro}}&= \boldsymbol{T_\text{head}^{\text{ext}}},\\
\displaystyle{\sum_{i=j}^N}\boldsymbol{T} _{i,\boldsymbol{X}^j}^{\text{hydro}}&= \boldsymbol{T}_{j}^{\text{ext}}, \quad \text{for } j=1,\ldots, N.
    \end{cases}
\end{equation}\\

\noindent {\bf Head force and torque:} The viscous force $\boldsymbol{F^{\text{hydro}}_{\text{head}}}$ and torque $\boldsymbol{T^{\text{hydro}}_{\text{head}}}$ acting on the head are computed using RFT. We have
\begin{align*}
    \boldsymbol{F^{\text{hydro}}_{\text{head}}} 
    &= r\left( -k^h_{\parallel}(\dot{\boldsymbol{X}} \cdot \boldsymbol{e}_1^h) \boldsymbol{e}_1^h
       - k^h_{\perp} (\dot{\boldsymbol{X}} \cdot \boldsymbol{e}_2^h) \boldsymbol{e}_2^h 
       - k^h_{\perp} (\dot{\boldsymbol{X}} \cdot \boldsymbol{e}_3^h) \boldsymbol{e}_3^h\right) \\
    &= -rR^h D^h (R^h)^T \dot{\boldsymbol{X}},
\end{align*}
where $D^h := \diag([k^h_{\parallel}, k^h_{\perp}, k^h_{\perp}]) \in \mathbb{R}^{3 \times 3}$ is a diagonal matrix, and $k^h_{\parallel}$ and $k^h_{\perp}$ are the parallel and perpendicular drag coefficients associated with the head. Due to the spherical shape of the head, we will consider that $k_{\parallel}^h=k_{\perp}^h$. The torque is given by
\begin{equation*}
    \boldsymbol{T^{\text{hydro}}_{\text{head}}} = - r^3k_r \boldsymbol{\Omega}^h,
\end{equation*}
where $k_r$ is the rotational drag coefficient for the spherical head, and $\boldsymbol{\Omega}^h \in \mathbb{R}^3$ is the angular velocity of the head corresponding to the rotation matrix $R^h$.\\

\noindent {\bf Links forces and torques :}
The density of the viscous force $\boldsymbol{f}_i^{\text{hydro}}$ acting on the $i$-th link is assumed to depend linearly on velocity. We therefore have, by RFT,
\begin{align*}
    \boldsymbol{f}_i^{\text{hydro}}(s) =&
    \begin{multlined}[t]
    -k_{\parallel}^i (\dot{\boldsymbol{x}}^i(s) \cdot \boldsymbol{e}_1^i)\boldsymbol{e}_1^i
    -k_{\perp}^i (\dot{\boldsymbol{x}}^i(s) \cdot \boldsymbol{e}_2^i)\boldsymbol{e}_2^i\\ 
    -k_{\perp}^i (\dot{\boldsymbol{x}}^i(s) \cdot \boldsymbol{e}_3^i)\boldsymbol{e}_3^i
    \end{multlined}\\
    =& -R^h \tilde{D}^i (R^h)^T  \dot{\boldsymbol{x}}^i(s),
\end{align*}
with $\tilde{D}^i := R^i D^i (R^i)^T \in \mathbb{R}^{3\times3}$ where $D^i := \diag([k_{\parallel}^i, k_{\perp}^i, k_{\perp}^i]) \in \mathbb{R}^{3\times3}$ is a diagonal matrix, $k^i_{\parallel}$ and $k^i_{\perp}$ are the parallel and perpendicular drag coefficients associated with the $i$-th link. Using \eqref{eq:dotxi}, we obtain that the force $\boldsymbol{F}_i^{\text{hydro}}$ is expressed as
\begin{align*}
    \boldsymbol{F}_i^{\text{hydro}} =& \int_{0}^l \boldsymbol{f}_i^{\text{hydro}}(s)ds\\
    =& 
    -\left(R^h \tilde{D}^i (R^h)^T\right)\\ &\left(l\dot{\boldsymbol{X}}^i+\frac{l^2}{2}[R^hR^i\boldsymbol{e}_1]^{\times}\boldsymbol{\Omega}^h+\frac{l^2}{2}R^h[R^i\boldsymbol{e}_1]^{\times}\boldsymbol{\Omega}^i\right).
\end{align*}
We denote by $\boldsymbol{T}^{\text{hydro}}_{i,\boldsymbol{x}^0}$ the $i$-th torque with respect to $\boldsymbol{x}^0$ such that
\begin{align*}
    \boldsymbol{T}^{\text{hydro}}_{i,\boldsymbol{x}^0} &= \int_{0}^l
(\boldsymbol{x}^i(s) - \boldsymbol{x}^0)\times \boldsymbol{f}_i^{\text{hydro}}(s) ds\\
&= (\boldsymbol{X}^i-\boldsymbol{x}^0)\times \boldsymbol{F}_i^{\text{hydro}} - (R^hR^i \boldsymbol{e}_1)\times \int_{0}^l s\boldsymbol{f}_i^{\text{hydro}}(s) ds\\
 &=\begin{multlined}[t]
      [\boldsymbol{X}^i-\boldsymbol{x}^0]^{\times}\boldsymbol{F}_i^{\text{hydro}} +[R^hR^i\boldsymbol{e}_1]^{\times}\left(R^h\tilde{D}^i(R^h)^T\right)\\   
      \left(\frac{l^2}{2}\dot{\boldsymbol{X}}^i + \frac{l^3}{3}[R^hR^i\boldsymbol{e}_1]^{\times}\boldsymbol{\Omega}^h + \frac{l^3}{3}R^h[R^i\boldsymbol{e}_1]^{\times}\boldsymbol{\Omega}^i\right).
    \end{multlined}
\end{align*}

\noindent {\bf External forces and torques :} The elasticity of the flagellum is modeled using discrete beam theory \cite{moreau2018}. Each joint between consecutive links is represented by a torsion spring with constant $k_{\text{el}}$. The elastic restoring torque at joint $\boldsymbol{X}^i$ is
\begin{equation*}
    \boldsymbol{T}^{\text{el}}_i = k_{\text{el}} \, \boldsymbol{e}_1^i \times \boldsymbol{e}_1^{i-1}.
\end{equation*}
The swimmer is actuated solely by an external magnetic torque. Let $\boldsymbol{M} = m \boldsymbol{e}_1^h$ denote the magnetic moment of the swimmer's head. For a homogeneous, time-dependent magnetic field $\boldsymbol{B}(t)=\begin{bmatrix}u_1(t), u_2(t), u_3(t)\end{bmatrix}^\top$, the torque applied to the swimmer is
\begin{equation*}
    \boldsymbol{T^{\text{mag}}} = \boldsymbol{M} \times \boldsymbol{B}.
\end{equation*}
Consequently, the external forces and torques in \eqref{eq:newtonsyst} are
$\boldsymbol{F_{\text{ext}}} = 0$, $\boldsymbol{T^{\text{ext}}_{\text{head}}} = - \boldsymbol{T^{\text{mag}}}$, and $\boldsymbol{T}^{\text{ext}}_{j} = - \boldsymbol{T}_j^{\text{el}}$.\\

\noindent {\bf Final system :}
Because of the parametrization of the vector $\boldsymbol{e}_1^i$, where rotation about this axis is negligible, only the projections of the torque onto the plane orthogonal to $\boldsymbol{e}_1^i$ are considered. This projection is denoted by $\Pi^{i}$ which is simply the second and third lines of $R^hR^i$. Consequently, the operator $\Pi^{j}$ must be applied to the last rows of \eqref{eq:newtonsyst}. Finally, the swimmer is described by two sets of variables:  the $2N+6$ state variables $(\boldsymbol{X}, \boldsymbol{\Theta}, \boldsymbol{\Phi})^\top$ such that for all $t\geq0$, 
\begin{align*}
    \boldsymbol{X}(t) &= (x(t),y(t),z(t)) \in \mathbb{R}^3,\\
    \boldsymbol{\Theta}(t) &= (\theta_x(t), \theta_y(t), \theta_z(t)) \in [0,2\pi]^3,\label{eq:state_nlinks}\\
     \boldsymbol{\Phi}(t)& = (\phi_y^1(t), \phi_z^1(t), \ldots, \phi_y^N(t), \phi_z^N(t)) \in [0,2\pi]^{2N},
\end{align*}
and the $3$ \textit{control variables} $\boldsymbol{B}$ such that
\begin{equation*}
    \boldsymbol{B}(t) = (u_1(t), u_2(t), u_3(t)) \in \mathbb{R}^3.
\end{equation*}
The system \eqref{eq:newtonsyst} can be written in the following matrix form
\begin{equation}\label{eq:mat_form}
AQB\begin{bmatrix}
        \dot{\boldsymbol{X}}\\
        \dot{\boldsymbol{\Theta}}\\
        \dot{\boldsymbol{\Phi}}
    \end{bmatrix}= \boldsymbol{F}_0 + u_1 \boldsymbol{F}_1 + u_2 \boldsymbol{F}_2 + u_3 \boldsymbol{F}_3.
\end{equation}
The general forms of the matrices $A, Q$ and $B$, as well as the vectors $\boldsymbol{F}_0$, $\boldsymbol{F}_1$, $\boldsymbol{F}_2$ and $\boldsymbol{F}_3$  are defined in Appendix~\ref{app:mat_form}

\subsection{Three-sphere swimmer model}

\subsubsection{Geometrical modeling}
\begin{figure}[htbp]
    \centering
    \includegraphics[width=1\linewidth]{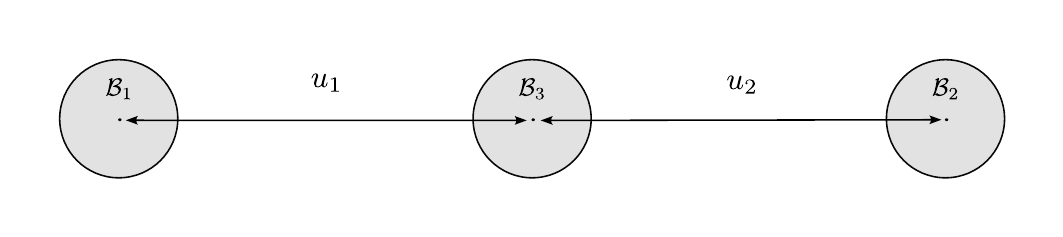}
    \caption{Illustration of the three-sphere swimmer. The left sphere is denoted by $\mathscr{B}_1$, the right sphere by $\mathscr{B}_2$, and the central reference sphere by $\mathscr{B}_3$. The swimmer propels itself by varying the arm lengths $u_L$ and $u_R$ at speeds $\dot u_1$ and $\dot u_2$, respectively.}
    \label{fig:three_sphere}
\end{figure}

The three-sphere swimmer introduced by Najafi and Golestanian \cite{najafi2004} consists of three collinear spheres of equal radius $R$ (see \Cref{fig:three_sphere}). The two lateral spheres, denoted by $\mathscr{B}_1$ and $\mathscr{B}_2$, are connected to the central sphere $\mathscr{B}_3$ by thin links, whose hydrodynamic effects are neglected. Throughout this work, the central sphere $\mathscr{B}_3$ is taken as the reference body for defining relative quantities.  Propulsion arises from time-dependent variations of the arm lengths $u_1$ and $u_2$, which are actuated at speeds $\dot u_1$ and $\dot u_2$, respectively. A non-reciprocal actuation sequence is required to break the time-reversal symmetry of Stokes equations and generate net motion.  

\subsubsection{Swimmer dynamics}
Let $\mathscr{F}^0 \subset \mathbb{R}^{d_p}$, $d_p=2$ in our study, but the model naturally extends to $d_p=3$, denote the fluid domain at initial time, and $\mathscr{F}^t$ its configuration at time $t$. 
As shown in Appendix~\ref{app:threesphere2Dvs3D}, the qualitative features of the swimmer’s displacement remain essentially unchanged when transitioning from two to three spatial dimensions. The $i$-th sphere occupies domain $\mathscr{B}_i^t \subset \mathbb{R}^{d_p}$, with translational velocity $\boldsymbol{V}_i$ and angular velocity $\boldsymbol{\omega}_i$. The fluid velocity and pressure fields are denoted by $\boldsymbol{v} : \mathscr{F}^t \to \mathbb{R}^{d_p}$ and $q : \mathscr{F}^t \to \mathbb{R}$, and $\mu$ is the dynamic viscosity.  The fluid–structure interaction is governed by the Stokes equations, coupled with rigid-body motion of the spheres:
\begin{equation}\label{eq:three_sphere_pde}
    \left\{
    \begin{aligned}
        -\mu \Delta \boldsymbol{v} + \boldsymbol{\nabla}q &= 0 & \quad &\text{in } \mathscr{F}^t,\\
        \boldsymbol{\nabla}\cdot \boldsymbol{v} &= 0 & \quad &\text{in } \mathscr{F}^t,\\
        \boldsymbol{v} &= \boldsymbol{V}_i + \boldsymbol{\omega}_i \times (\boldsymbol{x}-\boldsymbol{x}^{\text{CM}}) + \boldsymbol{v}_i^d & \quad &\text{on } \partial\mathscr{B}_i^t,\\ 
        m_i\dot{\boldsymbol{V}}_i &= -\boldsymbol{F}^{\text{fluid}} = 0_{\mathbb{R}^{d_p}}, & & \\
        J_i \dot{\boldsymbol{\omega}}_i &= -\boldsymbol{M}^{\text{fluid}} = 0_{\mathbb{R}^{d_p}}, & & 
    \end{aligned}
    \right.
\end{equation}
where $\boldsymbol{x}^{\text{CM}}$ is the center of mass of the swimmer, and $\sigma(\boldsymbol{v},q)=(\boldsymbol{\nabla}\boldsymbol{v}+\boldsymbol{\nabla}\boldsymbol{v}^\top)-qI_{d_p}$ is the Cauchy stress tensor. 
At the boundary of each sphere, the Dirichlet condition is decomposed into a rigid motion and a deformation velocity that represents the stroking strategy. The deformation velocities $(\boldsymbol{v}_1^d, \boldsymbol{v}_2^d, \boldsymbol{v}_3^d)$ represent the actuation. 
The hydrodynamic forces and torques are given by
\begin{align*}
\boldsymbol{F}^{\text{fluid}}(t) &= \sum_{i=1}^3 \int_{\partial \mathscr{B}_i^t}\sigma(\boldsymbol{v},q)\,\boldsymbol{n}\,ds\\
\boldsymbol{M}^{\text{fluid}}(t) &= \sum_{i=1}^3 \int_{\partial \mathscr{B}_i^t} (\boldsymbol{x}-\boldsymbol{x}^{\text{CM}}) \times \sigma(\boldsymbol{v},q)\,\boldsymbol{n}\,ds.
\end{align*}
At low Reynolds number, inertia is negligible, so $\boldsymbol{F}^{\text{fluid}} \equiv \boldsymbol{M}^{\text{fluid}} \equiv 0$. Since the spheres are connected by rigid arms, their relative velocities\footnote{In the case $d_p = 3$, the actuation involves rotations about the $-y$ and $-z$ axes, described using the corresponding rotation matrices.} are prescribed with respect to the body frame $\mathscr{B}_3$:
\begin{equation*}
\boldsymbol{V}_1 = \boldsymbol{V}_2 = \boldsymbol{V}_3,
\quad
\boldsymbol{v}_1^d =
\begin{bmatrix}
\cos(\theta)\\
\sin(\theta)
\end{bmatrix} \dot u_1, ~
\boldsymbol{v}_2^d =
\begin{bmatrix}
\cos(\theta)\\
\sin(\theta)
\end{bmatrix} \dot u_2, ~
\boldsymbol{v}_3^d \equiv 0,
\end{equation*}
where $\theta$ denotes the swimmer’s orientation in the laboratory frame. Equation \eqref{eq:three_sphere_pde} is solved using its weak formulation with $(\boldsymbol{v},q) \in (H^1(\mathscr{F}^t))^{d_p} \times L^2(\mathscr{F}^t)$ and $(\boldsymbol{V}_i,\boldsymbol{\omega}_i) \in (\mathbb{R}^{d_p})^3 \times (\mathbb{R}^{d_p})^3$. The finite element library \texttt{Feel++}\footnote{\url{https://docs.feelpp.org/feelpp/0.110/index.html}} is used for the discretization and numerical solution \cite{prudhomme2025, berti2022}.

\section{Optimal control problem}\label{sec:ocp}
In this section, the dynamics of the previously introduced swimmers are reformulated in a general form to write the trajectory tracking task as a constrained optimization problem, as defined in \eqref{eq:pb_opti_generic}. For each case, we specify the admissible control set according to physical constraints. Finally, we present the solution strategy, which combines B-splines for parametrization and Bayesian optimization for optimization.
\subsection{$N$-link model}

In the $N$-link model, the swimmer is actuated by an external magnetic field $\boldsymbol{B}$ applied to its magnetic head, which serves as the control input. The state vector is defined as
\begin{equation*}
\boldsymbol{p} = \begin{bmatrix}   \boldsymbol{X}&\boldsymbol{\Theta}&\boldsymbol{\Phi}\end{bmatrix}^\top,
\end{equation*}
where $d_p = 3$, $d_o = 3$, and $d_{\text{in}} = 2N + 3$ in \eqref{eq:swimmerstate}. The swimmer’s dynamics \eqref{eq:mat_form} can therefore be expressed in the following general form for all $t \in [0, T]$:
\begin{align*}
\boldsymbol{G}(\boldsymbol{p}(t), \boldsymbol{u}(t), t) 
=&~ \dot{\boldsymbol{p}}(t)
- \boldsymbol{F}_0(t)
+ u_1(t)\boldsymbol{F}_1(t)
+ u_2(t)\boldsymbol{F}_2(t)\\
+&~ u_3(t)\boldsymbol{F}_3(t)\\
=&~ 0_{\mathbb{R}^{2N+6}}.
\end{align*}
The constraint operator $\boldsymbol{C}$ reduces to the initial condition:
\begin{equation*}
    \boldsymbol{C}(\boldsymbol{p}(0)) = \boldsymbol{p}(0) - \boldsymbol{p}_0 = 0_{\mathbb{R}^{2N+6}}.
\end{equation*}
In the following, we set $\boldsymbol{p}_0 = 0_{\mathbb{R}^{2N+6}}$. The magnetic field is physically constrained and must remain bounded \cite{faris2020}. Accordingly, we introduce the following three admissible control sets:
\begin{subequations}\label{eq:adm_control_nlinks}
\begin{align}
    \mathscr{U}_{\text{ad}}^{y} 
    &:= \begin{multlined}[t]
        \Big\{ \boldsymbol{u} \in L([0,T],\mathbb{R}^3) \;\Big|\;
        u_1(t) = M,\; 
        \\|u_2(t)| \leq M, u_3(t) = 0,\; \forall t \in [0,T] \Big\}
       \end{multlined} \\
    \mathscr{U}_{\text{ad}}^{x,y} 
    &:= \begin{multlined}[t]
        \Big\{ \boldsymbol{u} \in L([0,T],\mathbb{R}^3) \;\Big|\;
        |u_1(t)| \leq M,\; \\|u_2(t)| \leq M, u_3(t) = 0,\; \forall t \in [0,T] \Big\}
       \end{multlined} \label{eq:adm_control_xy}\\
    \mathscr{U}_{\text{ad}}^{x,y,z}
    &:= \begin{multlined}[t]
        \Big\{ \boldsymbol{u} \in L([0,T],\mathbb{R}^3) \;\Big|\;
        |u_1(t)| \leq M,\;\\ |u_2(t)| \leq M, |u_3(t)| \leq M,\; \forall t \in [0,T] \Big\}.
       \end{multlined}
\end{align}
\end{subequations}
where $M$ denotes the uniform bound in the $\|\cdot\|_{\infty}$ norm of the magnetic field. In the following, we always set $M = 0.01$, as taken in \cite{faris2020, oulmas2019}.  The sets $\mathscr{U}_{\text{ad}}^{y}$ and $\mathscr{U}_{\text{ad}}^{x,y}$ constrain the swimmer to planar motion, provided that the initial condition is also planar. In contrast, the set $\mathscr{U}_{\text{ad}}^{x,y,z}$ allows for three-dimensional actuation, enabling the swimmer to deviate from the plane.

\subsection{Three-sphere swimmer model}
In the case of the three-sphere swimmer, locomotion is achieved by successively extending and retracting its two arms. For numerical reasons, particularly to avoid possible intersections between the spheres, the control variable is chosen to be the arm lengths rather than their velocities. The state vector is defined with respect to the central sphere as
\begin{equation*}
    \boldsymbol{p} =
    \begin{bmatrix}
        \boldsymbol{X}_3  & \theta & \boldsymbol{X}_1 & \boldsymbol{X}_2
    \end{bmatrix}^\top,
\end{equation*}
where $\boldsymbol{X}_1$, $\boldsymbol{X}_2$, and $\boldsymbol{X}_3$ denote the positions of the centers of the spheres $\mathscr{B}_1$, $\mathscr{B}_2$, and $\mathscr{B}_3$, respectively, and $\theta$ represents their common orientation. In this case, we have $d_p = 2$, $d_o = 1$, and $d_{\text{in}} = 4$ in \eqref{eq:swimmerstate}. We denote by $R(\theta)$ the rotation matrix in two dimensions, defined as 
\begin{equation*}
    R(\theta) =
    \begin{bmatrix}
        \cos(\theta) & \sin(\theta) \\
        -\sin(\theta) & \cos(\theta)
    \end{bmatrix}.
\end{equation*}
In the three-dimensional case, we must instead consider the vector 
$\boldsymbol{\theta} =
\begin{bmatrix}
    \theta_x & \theta_y & \theta_z
\end{bmatrix}^\top$
and the rotation matrices $R_x$, $R_y$, and $R_z$ as described in~\cite{berti2022}.\\

\noindent The governing operator $\boldsymbol{G}$ encodes the field equations and rigid-body dynamics:
\begin{equation*}
\boldsymbol{G}(\boldsymbol{p}, \boldsymbol{u}) =
\begin{bmatrix}
-\mu \Delta \boldsymbol{v} + \boldsymbol{\nabla}q\\[2pt]
\boldsymbol{\nabla}\cdot \boldsymbol{v}\\[2pt]
\ddot{\boldsymbol{p}}\\[2pt]
\dot{\boldsymbol{p}} - 
\begin{bmatrix}
\boldsymbol{V}_3 &
\omega_3 & \boldsymbol{V}_1&
\boldsymbol{V}_2 
\end{bmatrix}^\top
\end{bmatrix}
= 0_{\mathbb{R}^{3 d_p + 2 d_o + 2d_{\text{in}}+ 1}}.
\end{equation*}

\noindent The constraint operator $\boldsymbol{C}$ collects the boundary and initial conditions:
\begin{equation*}
\boldsymbol{C}_B(\boldsymbol{p}, \boldsymbol{u}) =
\begin{bmatrix}
\boldsymbol{v} - \left(\boldsymbol{V}_3 + \omega_3(\boldsymbol{x}-\boldsymbol{x}^{\mathrm{CM}}) + R(\theta) \dot{u}_1 \boldsymbol{e}_1\right)\\[3pt]
\boldsymbol{v} - \left(\boldsymbol{V}_3 + \omega_3(\boldsymbol{x}-\boldsymbol{x}^{\mathrm{CM}}) + R(\theta) \dot{u}_2 \boldsymbol{e}_1\right)\\[3pt]
\boldsymbol{v} - \left(\boldsymbol{V}_3 + \omega_3(\boldsymbol{x}-\boldsymbol{x}^{\mathrm{CM}})\right)
\end{bmatrix}
= 0_{\mathbb{R}^{3d_p}}
\end{equation*}
on the swimmer boundaries, and the initial condition
\begin{equation*}
\boldsymbol{C}_I(\boldsymbol{p}) = \boldsymbol{p}(0) - \boldsymbol{p}_0 = 0_{\mathbb{R}^{d_p+d_o+d_{\text{in}}}}.
\end{equation*}
Hence, $\boldsymbol{C} = (\boldsymbol{C}_B, \boldsymbol{C}_I,)$, 
where $\boldsymbol{C}_B$ represents the boundary coupling and $\boldsymbol{C}_I$ the initial data. In the following, we consider different values of $\boldsymbol{p}_0$, corresponding to varying initial heights.\\

\noindent Since the control inputs correspond to the lengths of the two arms, we define the admissible set of controls as
\begin{multline}\label{eq:adm_control_threesphere}
    \mathscr{U}_{\text{ad}}^{l,r} := \Big\{ \boldsymbol{u} \in L([0,T],\mathbb{R}^2) \;\Big|\; 
       m \leq u_1(t) \leq M,\;\\
       m \leq u_2(t) \leq M,\; \|\dot{\boldsymbol{u}}(t)\|_{\infty} \leq u'_{\text{max}},\; \forall t \in [0,T] \Big\},
\end{multline}
where $m$ and $M$ denote the minimum and maximum allowed lengths of the arms, and $u'_{\text{max}}$ is the maximum allowable velocity of arm deformation. Since the swimmer operates at the microscale, the rate of arm deformation is physically constrained. In the following, we set $m = 2R + R/4$ and $M = 10R$, where $R$ is the radius of each sphere, and $u'_{\text{max}} = 0.4$. These parameters allow us to analyze different regimes of the three-sphere swimmer. Depending on the spacing between the spheres, the hydrodynamic interactions can vary significantly \cite{golestanian2019}.

\subsection{Numerical solution}

\subsubsection{B-Splines}
 
 Let $\mathscr{T}=\{t_0,\ldots,t_n\}$ denote a nondecreasing sequence of real numbers, i.e., $t_i \leq t_{i+1}$ for $i=0,\ldots,n-1$. The values $t_i$ are called \textit{knots}, and $\mathscr{T}$ is referred to as the \textit{knot vector}. Several types of knot vectors exist, depending on the application \cite{piegl1997}. The $i$-th B-spline basis function of degree $d$ (also referred to as \textit{order} $d+1$), denoted $S_{i,d}$, is defined recursively as  
\begin{align*}
    S_{i,0}(t) &:= 
    \begin{cases}
        1 & \text{if } t \in [t_i,t_{i+1}), \\
        0 & \text{otherwise},
    \end{cases} \\
    S_{i,d}(t) &:= \frac{t-t_i}{t_{i+d}-t_i} S_{i,d-1}(t) + \frac{t_{i+d+1}-t}{t_{i+d+1}-t_{i+1}} S_{i+1,d-1}(t),
\end{align*}
with the convention that fractions with zero denominators are taken to be zero. Two important properties follow directly from the recursive definition: \textit{local support}, i.e  $S_{i,d}(t) = 0$ whenever $t \notin [t_i,t_{i+d+1})$ and \textit{partition of unity}, i.e  $\sum_{j=i-d}^i S_{j,d}(t) = 1$ for all $t \in [t_i,t_{i+1})$, with $S_{i,d}(t) \geq 0$ for all $t \in \mathbb{R}$.


\noindent A \textit{B-spline curve of degree $d$} is then defined by
\begin{equation*}
    S(t) := \sum_{i=0}^N S_{i,d}(t) P_i,
\end{equation*}
where $\mathscr{T}$ is a knot vector and the $P_i \in \mathbb{R}$ are the \textit{control points}. In this work, we focus on B-spline curves of dimension $1$ (i.e., $P_i \in \mathbb{R}$), but the construction extends directly to higher dimensions with $P_i \in \mathbb{R}^{\text{dim}}$.  In what follows, we consider knot vectors of the form
\begin{equation}\label{eq:knot_vector}
    \mathscr{T} = \{\underbrace{t_0,\ldots,t_0}_{d+1},\, t_1,\ldots,t_{n-1},\, \underbrace{t_n,\ldots,t_n}_{d+1}\},
\end{equation}
with equally spaced interior knots. The multiplicity of the first and last knots being $d+1$ ensures that the B-spline interpolates the first and last control points at $t=t_0$ and $t=t_n$. The degree $d$, the number of control points $N+1$, and the number of knots satisfy the relation $\card(\mathscr{T}) = N + d + 2.$\\

\begin{figure}[htbp]
    \centering
    \includegraphics[width=0.8\linewidth]{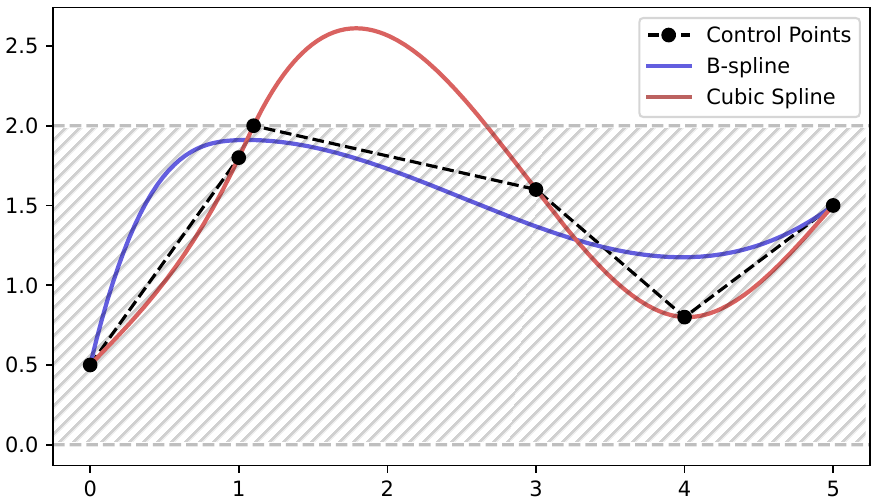}
    \caption{Illustration of the boundedness property of B-splines of degree $3$ with $\mathscr{T}=\{0, 0,0, 0, 1, 2, 5,5,5,5\}$ (respected) compared to cubic splines (not respected).}
    \label{fig:bspline_vs_cubic}
\end{figure}

\noindent Several spline bases could be used to parametrize the deformation. We employ B-splines since the resulting curve remains bounded by the extrema of its control points. This property, which does not hold for classical cubic splines, guarantees that admissibility constraints can be enforced simply by constraining the control points within the hypercube defined by the admissible set (see \eqref{eq:adm_control_nlinks_splines} and \eqref{eq:adm_control_threesphere_splines}). An illustration comparing B-splines and cubic splines is shown in \Cref{fig:bspline_vs_cubic}.  

\noindent To solve the optimal control problems under study, we parametrize the controls using one-dimensional B-spline curves. This yields the following admissible sets :
{\small
\begin{subequations}\label{eq:adm_control_nlinks_splines}
\begin{align}
    \tilde{\mathscr{U}}_{\text{ad}}^{y} 
    &:= \begin{multlined}[t]
     \Biggl\{ \boldsymbol{u} = \left(u_1, \sum_{i=0}^{N_u} S_{i,d} P^u_i, u_3\right) \;\Big|\; 
        u_1(t) = M,\;\\
        |P^u_i| \leq M,\; u_3(t) = 0,\; \forall t \in [0,T]  \Biggr\},     
    \end{multlined} \\
    \tilde{\mathscr{U}}_{\text{ad}}^{x,y} 
    &:= \begin{multlined}[t]
    \Biggl\{ \boldsymbol{u} = \left(\sum_{i=0}^{N_{u_1}} S_{i,d_1} P^{u_1}_i, \sum_{i=0}^{N_{u_2}} S_{i,d_2} P^{u_2}_i, u_3\right) \;\\\Big|\;
       |P^{u_1}_i| \leq M,\; |P^{u_2}_i| \leq M,\; u_3(t) = 0,\; \forall t \in [0,T] \Biggr\}, 
        \end{multlined}\\
   \tilde{\mathscr{U}}_{\text{ad}}^{x,y,z} 
    &:= \begin{multlined}[t]\Biggl\{ \boldsymbol{u} = \left(\sum_{i=0}^{N_{u_1}} S_{i,d_1} P^{u_1}_i, \sum_{i=0}^{N_{u_2}} S_{i,d_2} P^{u_2}_i, \sum_{i=0}^{N_{u_3}} S_{i,d_3} P^{u_3}_i\right) \;\\
    \Big|\; 
       |P^{u_1}_i| \leq M,\; |P^{u_2}_i| \leq M,\; |P^{u_3}_i| \leq M \Biggr\},
       \end{multlined}
\end{align}
\end{subequations}
}
and for the three-sphere swimmer,
\begin{multline}\label{eq:adm_control_threesphere_splines}
    \tilde{\mathscr{U}}_{\text{ad}}^{l,r} := \Biggl\{ \boldsymbol{u} = \left(\sum_{i=0}^{N_{u_1}} S_{i,d_1} P^{u_1}_i, \sum_{i=0}^{N_{u_2}} S_{i,d_2} P^{u_2}_i\right) \;\\
    \Big|\; 
       m \leq P_i^{u_1} \leq M,\; m \leq P_i^{u_2} \leq M,\; \\
       \|\dot{\boldsymbol{u}}(t)\|_{\infty} \leq u'_{\text{max}},\; \forall t \in [0,T] \Biggr\},
\end{multline}
which are finite-dimensional subsets of \eqref{eq:adm_control_nlinks} and \eqref{eq:adm_control_threesphere} due to the boundedness property of B-splines. Here, the natural numbers $d_1, d_2, d_3$ denote the degrees of the splines, and $N_{u_1}, N_{u_2}, N_{u_3}$ the numbers of control points associated with controls $u_1,u_2,u_3$, respectively. The number of control points has a significant impact on the results. A small number of control points restricts the admissible control space, limiting the class of representable functions and preventing good trajectory tracking. On the other hand, a huge number of control points increases the dimensionality of the optimization problem, making it more difficult to solve and leading to the presence of numerous local minima. In what follows, for the $N$-link swimmer we choose the number of control points equal to $40$, as it provides a good trade-off, as illustrated in Appendix~\ref{app:cost_vs_nbctrl}. For the three-sphere swimmer, due to numerical limitations inherent to the finite element computation of the Stokes dynamics, it was not possible to produce a similar convergence analysis similar. Consequently, the number of control points is fixed to $10$ for each control variable.\\

\subsubsection{Bayesian Optimization}

Bayesian optimization is a powerful approach for solving complex optimization problems where computing the gradient of the objective function is infeasible and evaluating the cost function is computationally expensive. The principle is to approximate the cost function with a surrogate model, typically a Gaussian Process (GP), built from a limited number of samples. This surrogate enables efficient predictions with negligible computational cost while also providing uncertainty estimates (see \cite{gramacy2020surrogates, Rasmussen2006}). \\

\noindent The \textit{Scalable Constrained Bayesian Optimization} (SCBO) algorithm \cite{eriksson_scalable_2021} extends this framework to large-scale constrained problems. It defines a trust region, represented by a hypercube, around the current best point. When constraints are not satisfied, the best point minimizes the maximum violation; otherwise, it minimizes the objective function. Unlike standard Bayesian optimization, SCBO selects a batch of candidate points using multiple realizations sampled from the GPs of the cost and constraint functions. The hypercube’s size and position are then updated based on the number of successful and failed evaluations. This method has been applied to shape optimization \cite{palazzolo2025a, hauke2024} and trajectory planning \cite{eriksson_scalable_2021}. To our knowledge, this is the first use of this method in the context of microswimmer trajectory tracking. The implementation is done by adapting the open-source SCBO code from \texttt{BoTorch}\footnote{\url{https://github.com/pytorch/botorch}}
 in \texttt{Python}. The SCBO parameters for all optimization problems are listed in \Cref{table:param_scbo}.

\section{Numerical Results}\label{sec:numres}

In \Cref{subsec:N-link}, trajectory tracking is carried out for the $N$-link swimmer along various reference paths, including both planar and non-planar trajectories. Long trajectories involving very high-dimensional optimization problems are also studied. \Cref{subsec:Three-spheres} is devoted to the three-sphere swimmer. In particular, the compensation of wall effects is studied across different regimes and altitude phases, where the presence of the wall induces angular drift through attraction or repulsion.

\subsection{$N$-links model}
\label{subsec:N-link}

\begin{figure}[htpb]
    \centering
    \includegraphics[width=1\linewidth]{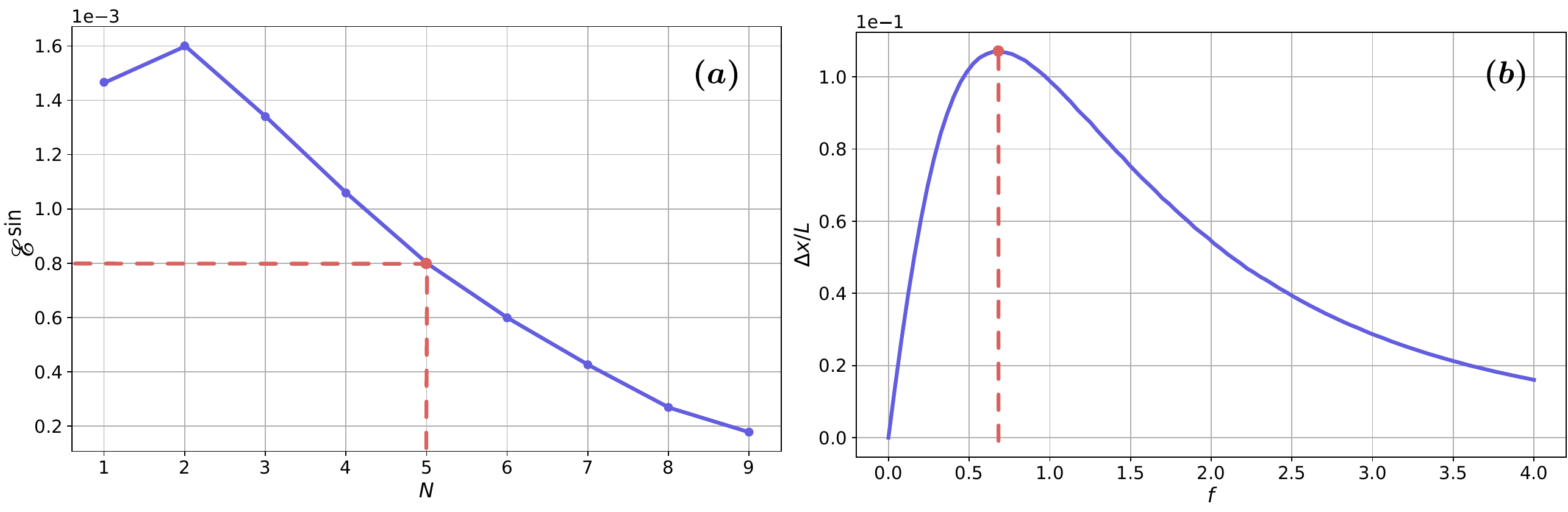}
   \caption{\textbf{(a)} Trajectory error in the $L^2$ norm between an $N$-link swimmer and an $(N+1)$-link swimmer under the magnetic field \eqref{eq:sin_ctrl}, with $B=0.01$, $f=0.5$, and $T=4/f$. \textbf{(b)} Mean displacement $\Delta x$ per period for various frequencies using the magnetic field defined in \eqref{eq:sin_ctrl}.}
    \label{fig:errors_N_Np1_and_disp_freq_sin}
\end{figure}

The $N$-links model, coupled with RFT, is parameterized by the discretization level of the fiber, specifically the number of links $N$. In \Cref{fig:errors_N_Np1_and_disp_freq_sin}, we display the trajectory error in the $L^2$ norm, between swimmers composed of $N$ and $N+1$ links, defined as
\begin{equation*}
    \mathscr{E}^{\text{sin}}(N) :=\left\|\boldsymbol{X}^{(N+1)}(\boldsymbol{u}_{\text{sin}}) - \boldsymbol{X}^{(N)}(\boldsymbol{u}_{\text{sin}})\right\|_{L^2([0,T];\,\mathbb{R}^3)},
\end{equation*}
where the control input is a planar sinusoidal magnetic field given by
\begin{equation}\label{eq:sin_ctrl}
    \boldsymbol{u}_{\text{sin}}(t) = B\begin{bmatrix}1\\ \sin(2\pi f t)\\ 0\end{bmatrix},
\end{equation}
where the amplitude is set to $B = 0.01$, the frequency to $f = 0.5$, and the final time to $T = 4/f$. The results show that choosing $N=5$ offers a good trade-off between accuracy and computational cost, yielding an $L^2$ error of less than $10^{-3}$ over four oscillation periods. Consequently, we fix $N=5$ for all subsequent simulations unless specified otherwise. A classical strategy for trajectory tracking involves applying a magnetic field aligned with the tangent to the reference trajectory, combined with an oscillating component perpendicular to it \cite{alouges2015, Dreyfus2005, oulmas2017}. In \Cref{fig:errors_N_Np1_and_disp_freq_sin}, we report the displacement $\Delta x$ per period as a function of the frequency $f$. We observe that the optimal frequency is approximately $f_{\text{opt}} \approx 0.7$ Hz, consistent with previous findings in \cite{oulmas2017, alouges2015}.

\noindent In the following, we investigate the use of Bayesian optimization combined with B-spline-based controls to enhance trajectory tracking performance.\\

\subsubsection{Planar trajectories}
We begin by considering the planar case.\\

\noindent {\bf Maximal distance :}  The final time is set to $T=3$, and we aim to maximize the displacement along the $x$-axis.  This leads to the following optimization problems :
\begin{equation}\label{eq:pb_cible_2D}
     \inf_{\boldsymbol{u}\in \tilde{\mathscr{U}}_{\text{ad}}^{y}} -\frac{X_1(\boldsymbol{u},T)}{L},\quad \text{and} \quad \inf_{\boldsymbol{u}\in \tilde{\mathscr{U}}_{\text{ad}}^{x,y}} -\frac{X_1(\boldsymbol{u},T)}{L},
\end{equation}
where $\tilde{\mathscr{U}}_{\text{ad}}^{y}$ and $\tilde{\mathscr{U}}_{\text{ad}}^{x,y}$ denote the sets of admissible controls restricted to motions in the $x$–$y$ plane (i.e., trajectories lying in the plane $z=0$), as defined in \eqref{eq:adm_control_nlinks_splines}. This problem is equivalent to \eqref{eq:cost_func} with a distant final state and weighting matrices $Q = 0_{2N+6,2N+6}$ and $S = \operatorname{diag}([-1, 0,\ldots, 0])$. Each control spline is parameterized by $40$ control points ($N^{u_1}=N^{u_2}=40$), which results in an optimization problem of dimension $40 \times \text{(number of controls to be optimized)}$.\\

\begin{figure}[htpb]
    \centering
    \includegraphics[width=1\linewidth]{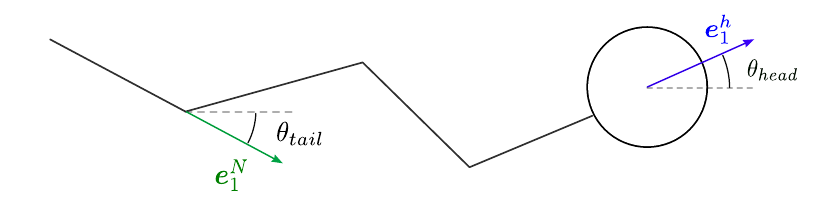}
\caption{Illustration of the angles $\theta_{\text{head}}$ and $\theta_{\text{tail}}$ in the $N$-links model, used to characterize the geometrical deformation of the swimming stroke in the planar case ($z=0$ plane).}
    \label{fig:theta_head_theta_tail}
\end{figure}

\noindent 
Classically, in the study of microswimming, the swimmer’s stroke is characterized through a phase portrait. In our framework, the phase portrait is defined by  $(\theta_{\text{tail}}, \theta_{\text{head}})$, where the head and tail orientations are given by $\theta_{\text{head}} = \theta_3$ and $\theta_{\text{tail}} = \theta_{\text{head}} + \phi_z^{N}$, as illustrated in \Cref{fig:theta_head_theta_tail}. The variables $\theta_3$ and $\phi_z^{N}$ are introduced in \Cref{sec:geo_nlinks}.

\begin{figure}[htpb]
    \centering
    \includegraphics[width=1\linewidth]{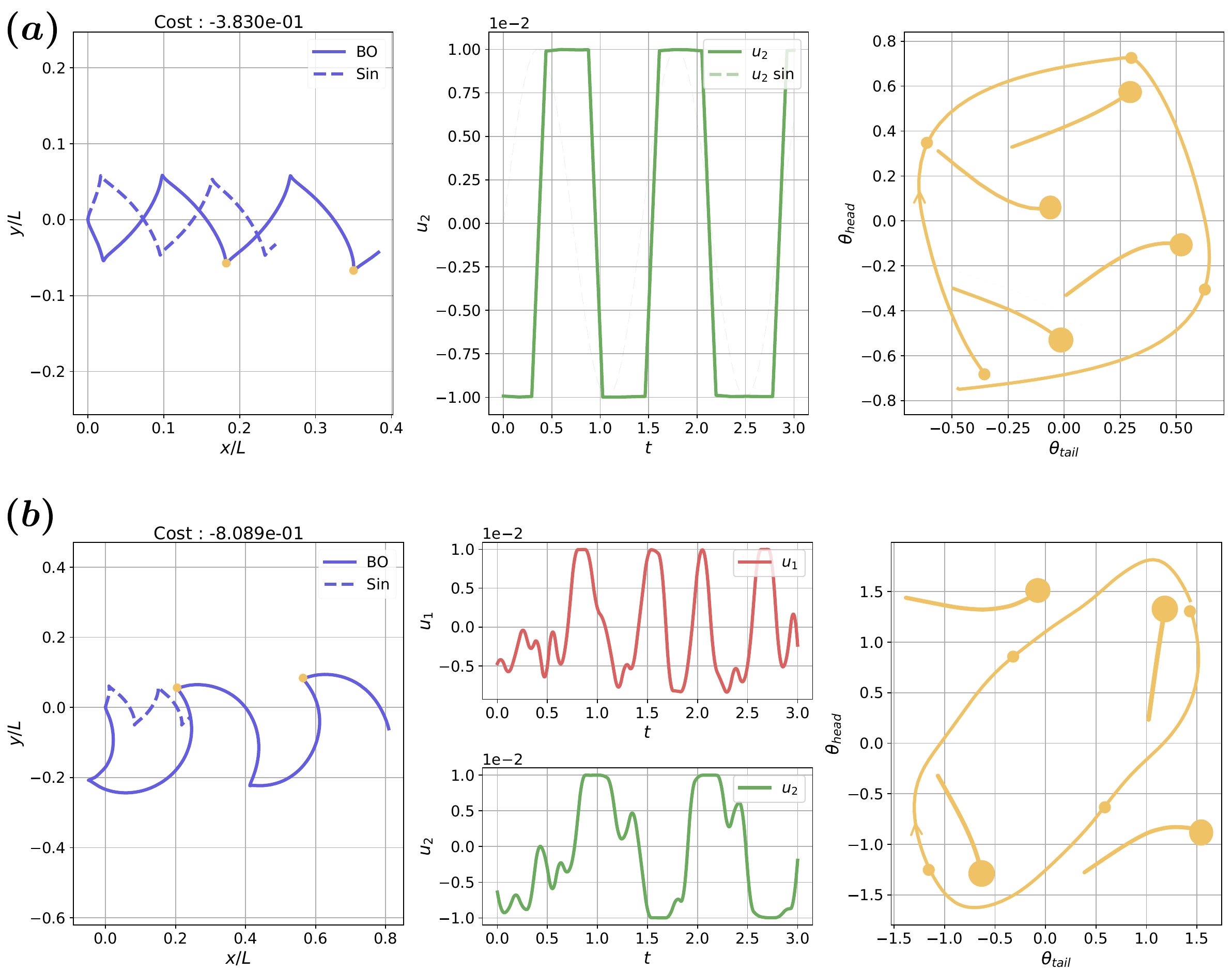}
\caption{Optimal trajectories for maximal displacement. The optimized trajectory (solid line, right panel) is compared with the reference trajectory under the sinusoidal control from \eqref{eq:sin_ctrl} (dashed line). Optimized magnetic controls $u_1$ and $u_2$ are shown in the middle panel (red and green, respectively). Start and end points of the stroke are highlighted in yellow. The phase portrait of $\theta_{\text{tail}} - \theta_{\text{head}}$ and swimmer deformation is shown in yellow. \textbf{(a)} Optimization within $\tilde{\mathscr{U}}_{\text{ad}}^y$, where only the $y$-component is optimized. \textbf{(b)} Optimization within $\tilde{\mathscr{U}}_{\text{ad}}^{x,y}$, where both $x$- and $y$-components are optimized.}
\label{fig:N_link_cible}
\end{figure}

\noindent As shown in \textbf{(a)} of \Cref{fig:N_link_cible}, only the $y$-component of the external magnetic field is optimized. Compared to the trajectory obtained under sinusoidal actuation at the optimal frequency $f_{\text{opt}}$, the Bayesian-optimized control performs significantly better. The resulting trajectory is nearly periodic, as the corresponding control $u_2$. Interestingly, the optimal control is not sinusoidal but rather of \textit{bang-bang} type, a well-known feature in optimal control theory \cite{pontryagin1987, trelat2005}. In the trajectory plot on the left, the amplitude of motion in the $y$-direction remains essentially unchanged, whereas the optimized swimmer exhibits a larger displacement in the $x$-direction.

In \textbf{(b)}, both components of the external magnetic field are optimized. The resulting trajectory outperforms the previous case and clearly surpasses the sinusoidal-field scenario. In the trajectory plot on the left, the motion amplitude is significantly larger, which, due to the flagellum’s elasticity, generates stronger propulsion impulses. Once again, the trajectory remains nearly periodic, while the corresponding phase portrait becomes more elongated. Notably, the swimmer reaches an almost vertical configuration at certain instants.\\

\noindent {\bf Portion of elliptical trajectory :}  We next perform a series of trajectory tracking tasks for elliptical reference paths of length $L$, which is close to the distance traveled for the problem \eqref{eq:pb_cible_2D} with $\tilde{\mathscr{U}}_{\text{ad}}^{x,y}$, beginning in the origin. The reference trajectory is defined by
\begin{equation}\label{eq:ellipse_traj}
\boldsymbol{p}_{\text{ref}}(t) = \begin{bmatrix}a + a \cos(-\frac{t}{T} s_{\text{end}}+\pi)\\ b \sin(-\frac{t}{T}s_{\text{end}}+\pi)\\ 0_{\mathbb{R}^{2N+4}}\end{bmatrix},
\end{equation}
with $L = \int_{0}^{s_{\text{end}}}\sqrt{a^2\sin^2(s) + b^2\cos^2(s)}\, ds$ where $a$ and $b$ are the semi-axes of the ellipse, and $s_{\text{end}}$ is computed via a bisection method to satisfy the arc-length condition. We set $a =L$ and vary $b$ among $\{\frac{L}{2}, L, \frac{3L}{2}\}$, to explore different curvature configurations. We set the final time at $T=3$ and we take $40$ number of control points for each spline. The cost function \eqref{eq:cost_func} is studied with $Q=10^9\diag([1,1, 0,\ldots, 0])$ and $S=10^4\diag([1,1,0,\ldots,0])$.\\

\begin{figure}[htpb]
    \centering
    \includegraphics[width=1\linewidth]{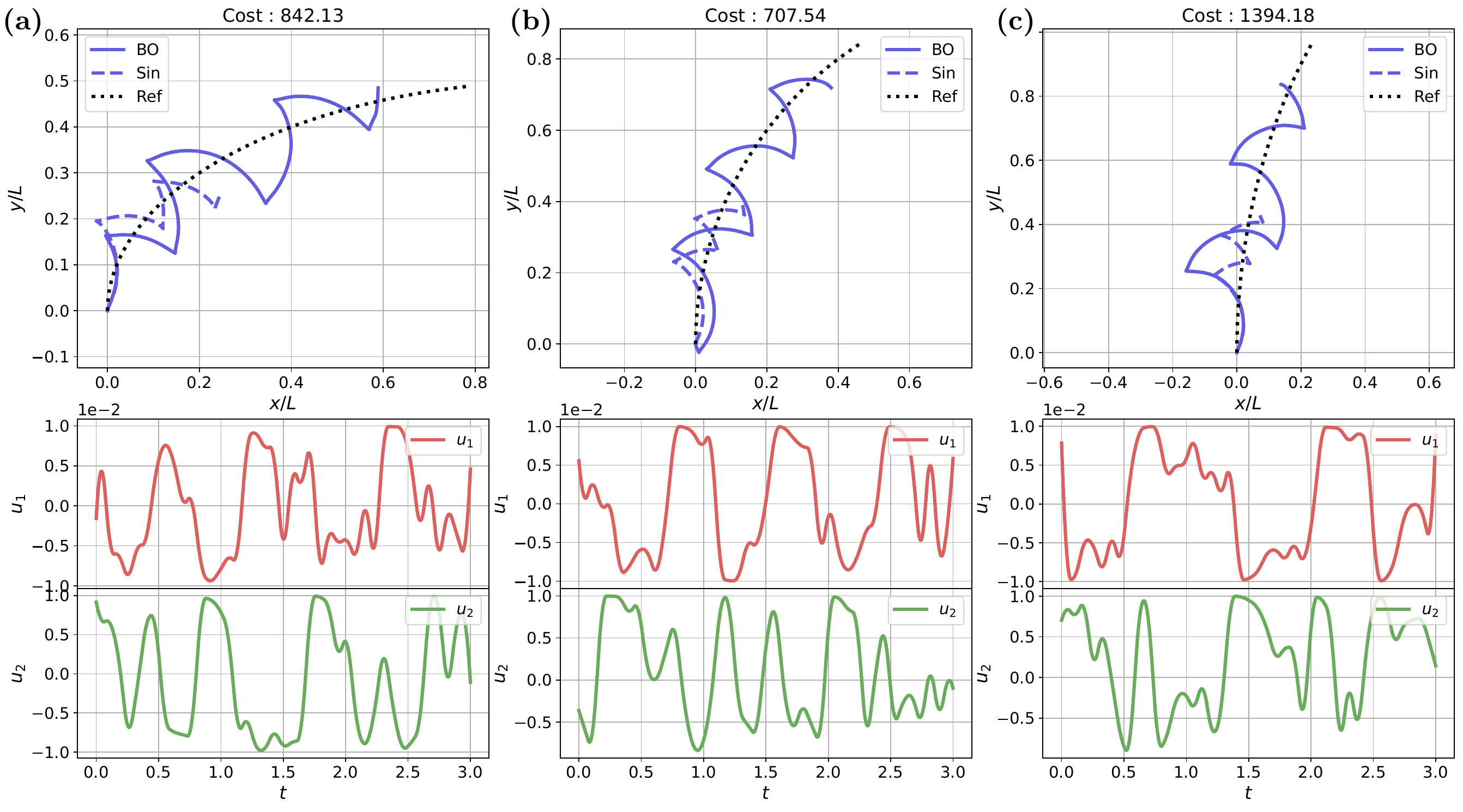}
    \caption{Trajectories for portion of elliptical references. From left to right: results for $b = \frac{L}{2}$, $b = L$, and $b = \frac{3L}{2}$. Top row: optimal trajectory (blue), trajectory under sinusoidal magnetic field \eqref{eq:sin_ctrl} with tangent alignment (dashed line), and reference trajectory \eqref{eq:ellipse_traj} (black). Bottom row: optimized controls $u_1$ (red) and $u_2$ (green).}
    \label{fig:N_link_2D_ellipses}
\end{figure}

\noindent The resulting optimal trajectories are compared with those obtained under the classical control at $f_{\text{opt}}$. \Cref{fig:N_link_2D_ellipses} shows that the optimized trajectories clearly outperform the classical ones. The optimal controls naturally adapt to the curvature of each reference path, highlighting the benefits of flexibility in control design. Interestingly, the resulting trajectories seems to be periodic. In this case, the phase portraits are not shown, as the curved paths make it difficult to define them as closed path.
\\

\noindent {\bf Complete elliptical trajectory :} In the following, we attempt to track elliptical trajectories of varying radii. We consider two distinct ellipses with radii chosen from the set $(a,b)\in\{\frac{L}{2}\}\times\{\frac{L}{2}, L\}$ as \eqref{eq:ellipse_traj}. One of the primary challenges is that as the length of the trajectory increases, both the final time and the number of control points must be correspondingly increased to ensure accurate tracking. For this study, we fix the final time at $T=10$ (respectively $T=15$) for $b=\frac{L}{2}$ (respectively $b=L$) and set the number of control points to $N^{u_1}=N^{u_2} = 100$. So the dimension of the optimization problem is equal to $200$. The cost function \eqref{eq:cost_func} is studied with $Q=10^9\diag([1,1, 0,\ldots,0])$ and $S=10^4\diag([1,1,0,\ldots,0])$.\\

\begin{figure}[htpb]
    \centering
    \includegraphics[width=1\linewidth]{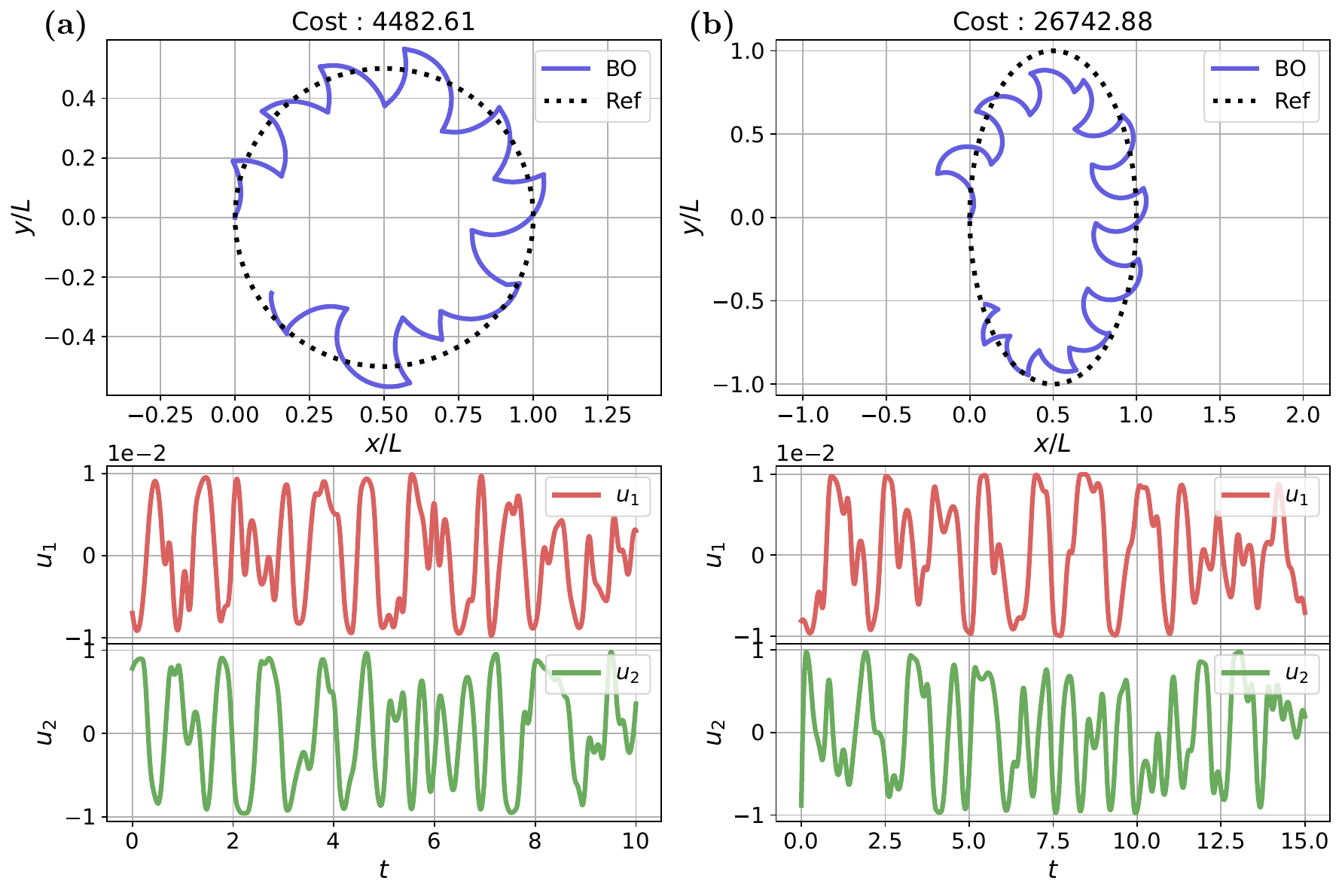}
\caption{Trajectories for elliptical references. 
\textbf{(a)} Case $a=b=\tfrac{L}{2}$. 
\textbf{(b)} Case $a=\tfrac{L}{2}, \, b=L$. 
Top row: optimal trajectory (blue) compared with the reference trajectory (black). 
Bottom row: optimized controls $u_1$ (red) and $u_2$ (green).}
\label{fig:N_link_2D_complete_ellipses}
\end{figure}

\noindent As shown in \Cref{fig:N_link_2D_complete_ellipses}, the microswimmer follows the reference trajectory with good accuracy. In both cases, characteristic optimal periodic motion patterns, consistent with those observed in previous simulations, reappear. Furthermore, the resulting trajectory resembles the path of a real sperm cell described in \cite{friedrich2010}. The small deviation from the target endpoint results from two main factors. First, increasing the terminal-state penalty weight in the cost function could improve endpoint accuracy but would degrade intermediate tracking performance. Second, as previously discussed, the optimization problem is highly sensitive to the choice of final time and the number of spline control points (see Appendix \ref{app:cost_vs_nbctrl}). Extending the time horizon or increasing the spline resolution could therefore provide greater flexibility, enabling more accurate endpoint convergence.\\

\subsubsection{Non-planar trajectories}

A series of trajectory-tracking simulations are performed for ellipsoidal path segments of arc length $L$, starting from the origin, using the admissible control set $\tilde{\mathscr{U}}_{\text{ad}}^{x,y,z}$. The reference trajectories are defined by
\begin{equation}\label{eq:ellipsoid_traj}
\boldsymbol{p}_{\text{ref}}(t) =
\begin{bmatrix}
a - a \cos^2(\frac{t}{T} s_{\text{end}})\\
b \cos(\frac{t}{T} s_{\text{end}})\sin(\frac{t}{T} s_{\text{end}})\\
c\sin(\frac{t}{T} s_{\text{end}})\\
0_{\mathbb{R}^{2N+3}}
\end{bmatrix},
\end{equation}
where the arc length $L$ is given by
{\footnotesize
$$
L = \int_{0}^{s_{\text{end}}}\sqrt{
4a^2\cos^2(s)\sin^2(s) + b^2(\cos^2(s) - \sin^2(s))^2 + c^2 \cos^2(s)
}\, ds,
$$
}with $a$, $b$, and $c$ denoting the semi-axes of the ellipsoid. The parameter $s_{\text{end}}$ is computed via a bisection method to ensure the prescribed arc length condition. In our study, we take $a = b = c = \frac{L}{2}$ and compare results obtained for two final times, $T \in \{3, 6\}$. Each control is parameterized using $40$ control points. The cost function \eqref{eq:cost_func} is used with weighting matrices $Q=10^9\diag([1,1,1, 0,\ldots,0])$ and $S=10^4\diag([1,1,1,0,\ldots,0])$.\\

\begin{figure}[htpb]
    \centering
    \includegraphics[width=1\linewidth]{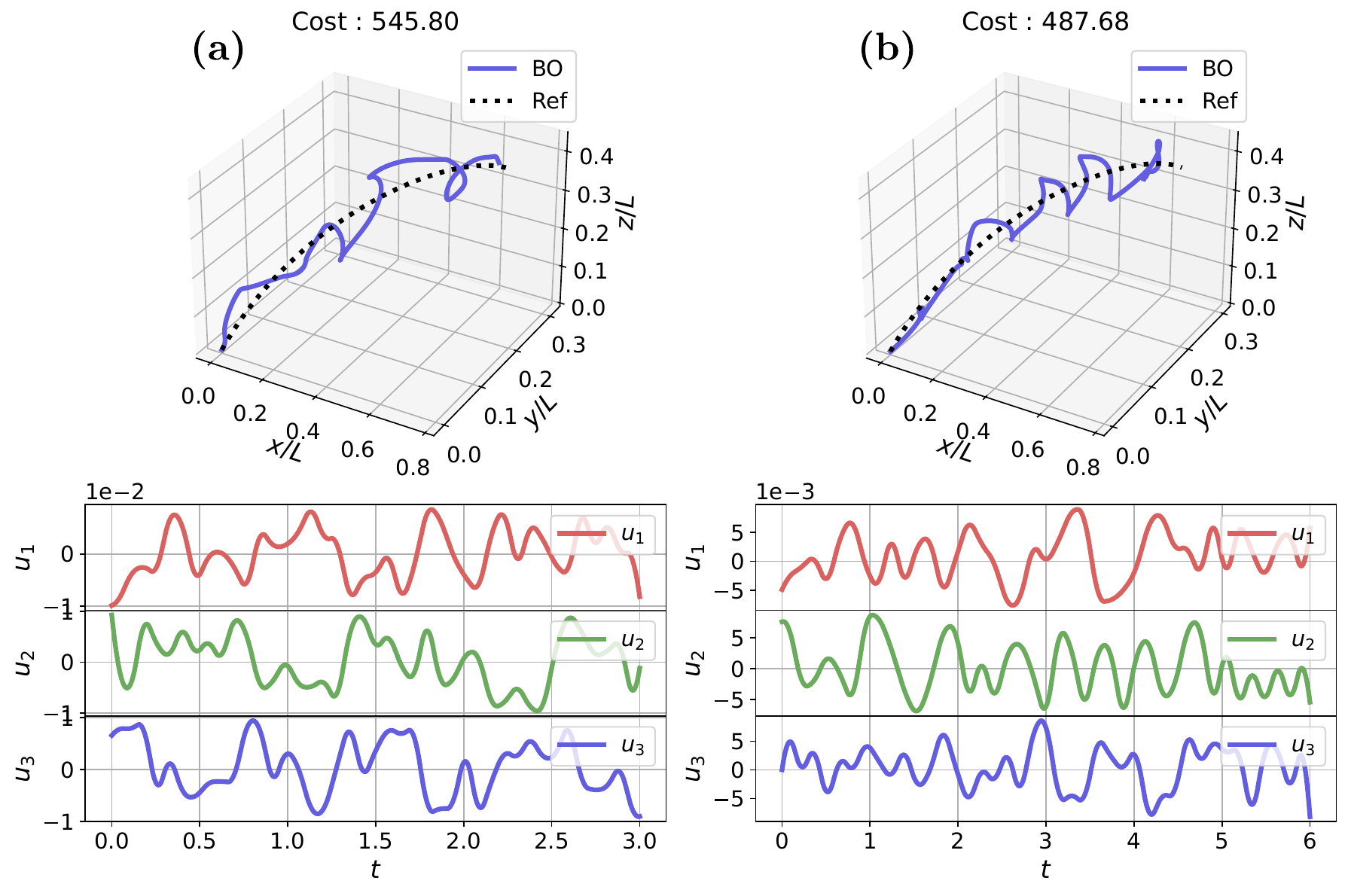}
\caption{Trajectories for ellipsoidal references. 
\textbf{(a)} Case $T=3$. 
\textbf{(b)} Case $T=6$. 
Top row: optimal trajectory compared with the reference trajectory (black). 
Bottom row: optimized controls $u_1$ (red), $u_2$ (green) and $u_3$ (blue).}
\label{fig:N_link_3D_ellipsoid}
\end{figure}

\noindent The resulting trajectories are shown in \Cref{fig:N_link_3D_ellipsoid}. The swimmer tracks the reference trajectory with high accuracy. However, unlike the planar case, no clear periodic patterns are observed in the trajectories.
However, the resulting trajectories rotate around the prescribed curve, as observed in experimental studies of real sperm cells \cite{jikeli2015}.
Additionally, increasing the final time $T$ allows the swimmer to follow the reference path more smoothly: for $T = 6$ (right panel in \Cref{fig:N_link_3D_ellipsoid}), the swimmer performs broader rotational motions around the reference path, whereas for $T = 3$ (left panel), the motion is more constrained as the swimmer must "catch up" more rapidly with the target point along the path.

\subsection{Three-sphere swimmer}
\label{subsec:Three-spheres}
One of the main challenges in the study of the three-sphere swimmer lies in determining the optimal periodic strokes that enable efficient propulsion through the fluid. To facilitate the enforcement of this periodicity constraint, we exploit the properties of B-spline interpolation at the first and last control points (a consequence of the chosen knot vector \eqref{eq:knot_vector}). Specifically, we impose $P_{1}^{u_1}=P_{N_{u_1}}^{u_1}$ and $P_{1}^{u_2}=P_{N_{u_2}}^{u_2}$ in \eqref{eq:adm_control_threesphere_splines}, which directly ensures $u_1(0)=u_1(T)$ and $u_2(0)=u_2(T)$. We will consider one periodic strokes of $T=4$, $N^{u_1}=N^{u_2}=10$ and $d=2$. \\

\begin{figure}[htbp]
    \centering
    \includegraphics[width=1\linewidth]{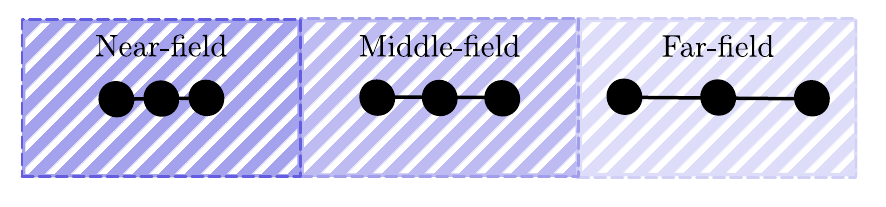}
    \caption{Illustration of the three different fields according the arm distances.}
    \label{fig:fields_3spheres}
\end{figure}

The initial condition places the central sphere at the center of a box of length $48R$ and height $80R$, with initial arm distances chosen to cover three hydrodynamic regimes (see \Cref{fig:fields_3spheres}):
\begin{itemize}
    \item \textit{Near-field}, with arm distances $2R+R/2$,
    \item \textit{Middle-field}, with arm distances $5R$,
    \item \textit{Far-field}, with arm distances $10R$.
\end{itemize}
The distances between the spheres define three distinct hydrodynamic regimes, characterized by varying intensities of lubrication effects: strong when the spheres are close, moderate at intermediate distances, and negligible in the \textit{far-field} regime (see, for instance, \cite{golestanian2019}, which discusses the impact of different hydrodynamic regimes on optimal control problems). While the control bounds in \eqref{eq:adm_control_threesphere} allow the regime to change during optimization, the initial and, by periodicity, the final conditions remain fixed.\\

\subsubsection{Optimal stroke}
\begin{figure}[htpb]
    \centering
    \includegraphics[width=1\linewidth]{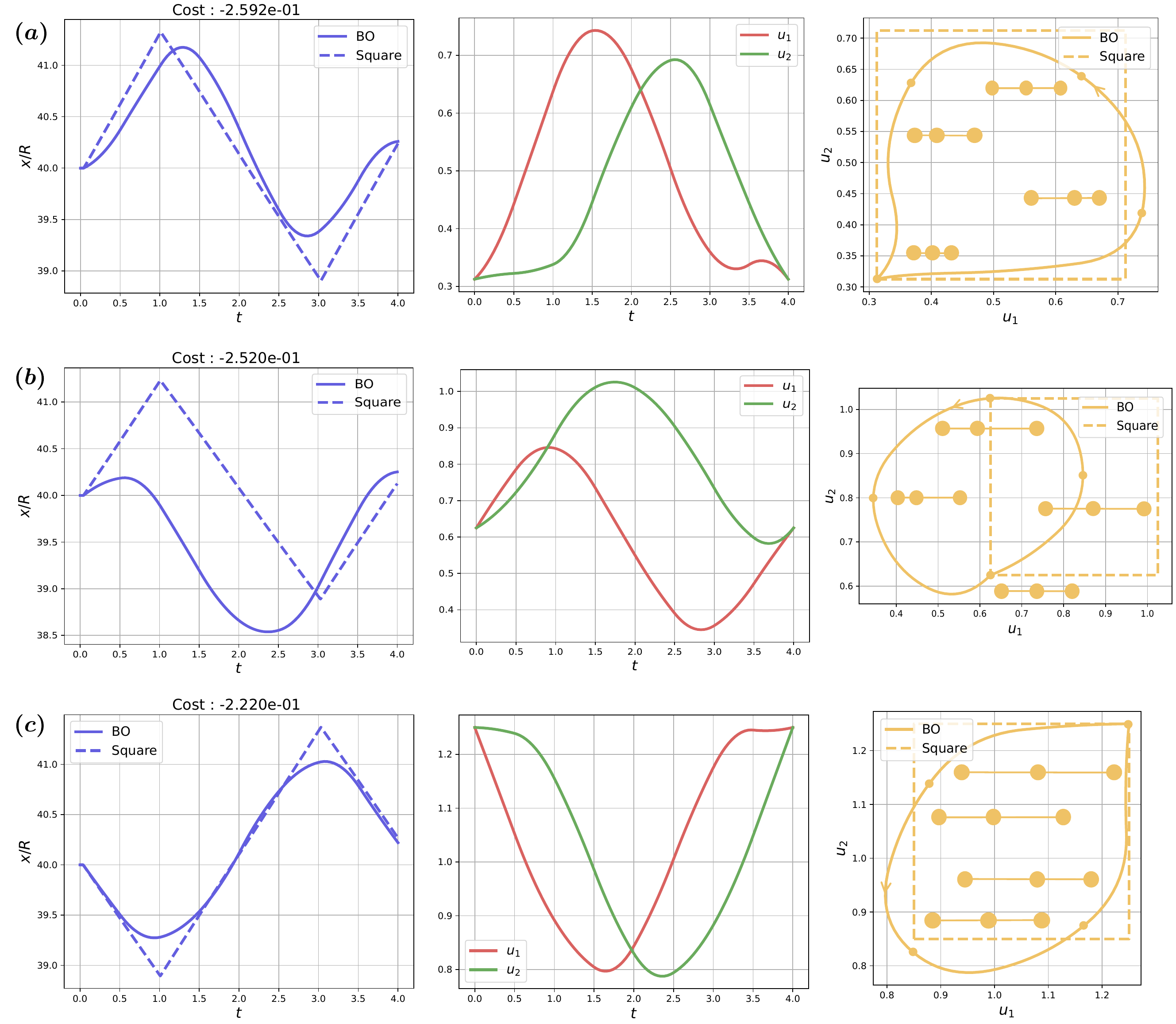}
\caption{Trajectories and control inputs of the optimized swimmer (solid lines) and the classical one (dotted lines). The blue curve shows the $x$-displacement over time, while the red and green curves correspond to the optimized controls $u_1$ and $u_2$, respectively. The yellow plot depicts the optimized and classical controls in the phase portrait. Optimized deformation patterns are also illustrated. \textbf{(a)} \textit{Near-field} regime with initial arm length $2R + R/2$. \textbf{(b)} \textit{Middle-field} regime with initial arm length $5R$. \textbf{(c)} \textit{Far-field} regime with initial arm length $10R$.}
\label{fig:three_sphere_max_dist}
\end{figure}

The classical stroke of the three-sphere swimmer consists of four sequential deformations, during which the two arms alternately contract and extend. This alternation ensures a non-reciprocal motion  \cite{najafi2004, golestanian2019, elKhiyati2023}. We compare the optimal stroke obtained through the proposed optimization framework with this classical one. For this purpose, a swimmer located sufficiently far from any boundaries is considered. The following cost functional is then minimized:
\begin{equation}\label{eq:max_dist_three_sphere}
\inf_{\boldsymbol{u} \in \tilde{\mathscr{U}}^{l,r}} -\frac{\left( X_1(\boldsymbol{u}, T)-X_1(\boldsymbol{u},0)\right)}{R}.
\end{equation}
\Cref{fig:three_sphere_max_dist} illustrates the resulting trajectories and corresponding optimal controls obtained from \eqref{eq:max_dist_three_sphere} for the three regimes. The optimized swimmer achieves a final displacement that is generally slightly higher than that of the classical stroke. For both the \textit{near-} and \textit{far-field} regimes, the optimal controls are very similar, except that one appears to be approximately the reversed deformation of the other. The points in the phase portraits that appear non-differentiable are due to the enforcement of the periodicity constraint and the fact that the two arms should begin and end with the same length.\\

\subsubsection{Impact of wall on the optimal stroke}
\begin{figure}[htbp]
    \centering
\includegraphics[width=1\linewidth]{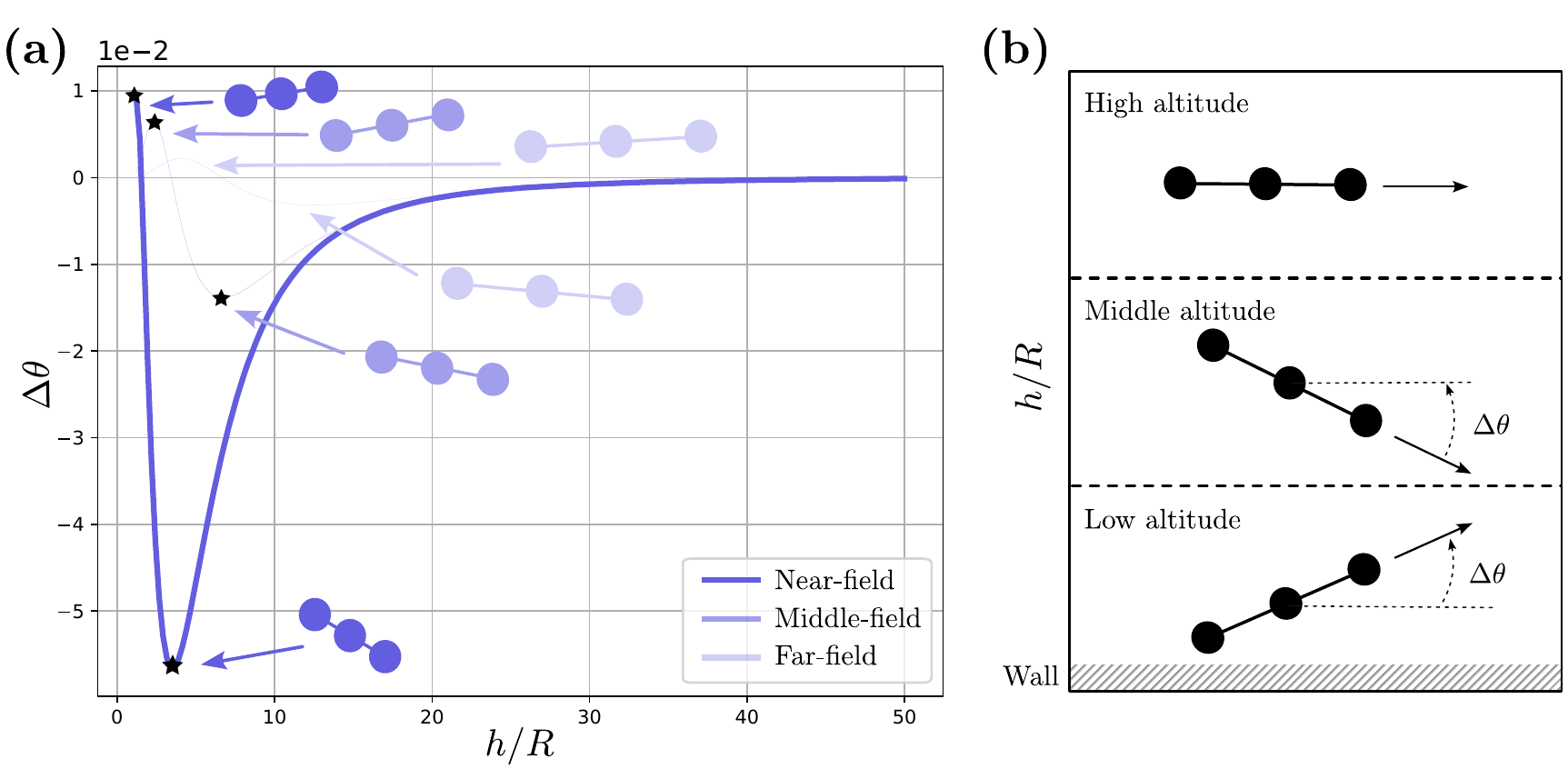}
\caption{\textbf{(a)} Angular deviation $\Delta \theta = \theta(T)-\theta(0)$ of the swimmer’s orientation after a single stroke at different heights $h$ from the wall. The three curves represent the behaviour of the \textit{near-}, \textit{middle-}, and \textit{far-field} regimes (from dark blue to light blue). For each regime, the swimmer’s orientation corresponding to the largest deviation observed in the first two phases is also shown. Black stars represent the cases studied in the wall–effects compensation optimization. \textbf{(b)} General behaviour of swimmers in each regime as a function of distance from the wall. The three phases are \textit{low-altitude}, \textit{middle-altitude}, and \textit{high-altitude}, corresponding respectively to repulsion from the wall, attraction to the wall, and no deviation.}
\label{fig:impact_of_wall_three_sphere}
\end{figure}

Taking the presence of walls into account when determining the optimal stroke remains a challenging problem. \Cref{fig:impact_of_wall_three_sphere} shows the angular deviation $\Delta \theta = \theta(T) - \theta(0)$ of the swimmer’s orientation over a single stroke for various distances $h$ from the wall, across the three different regimes (with $\theta(0)=0$). Three distinct phases can be observed with respect to the wall distance: 
\begin{itemize}
    \item \textit{Low-altitude}: close to the wall, the swimmer tends to move away from it.
    \item \textit{Middle-altitude}: at intermediate distances, the swimmer tends to approach the wall.
    \item \textit{High-altitude}: far from the wall, no significant deviation occurs.
\end{itemize}
The transition heights between these phases depend on the hydrodynamic regime considered. The closer the regime is to the \textit{near-field}, the larger the angular deviations across the phases. For $h = 20R$, all regimes exhibit similar behaviour.\\

\subsubsection{Wall effects compensation}
\begin{figure*}[htpb]
    \centering
    \includegraphics[width=1\linewidth]{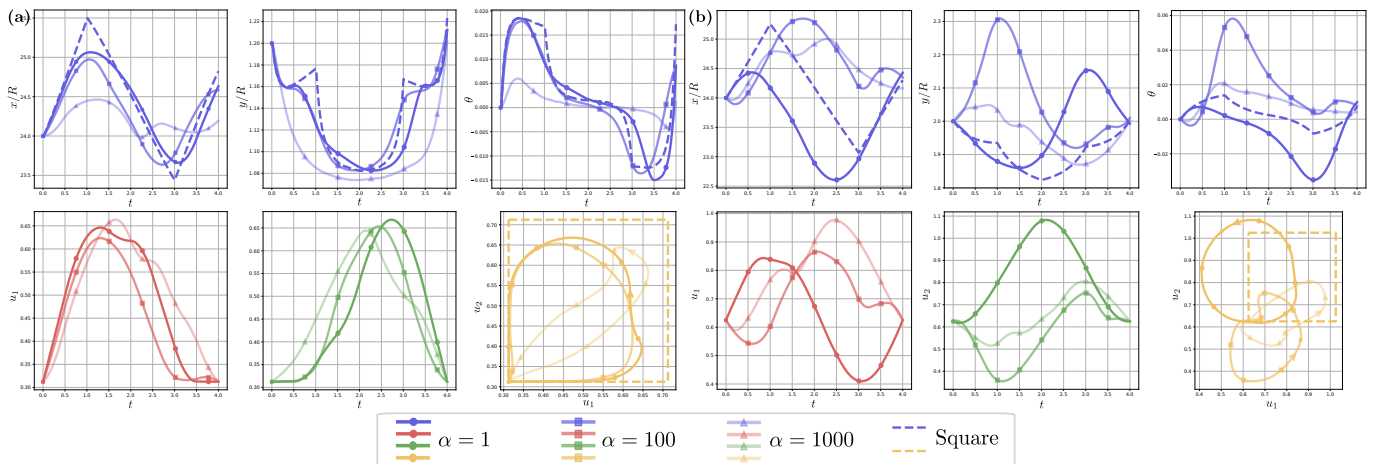}
\caption{Trajectories and control inputs of the optimized swimmers (solid lines) and the classical swimmer (dotted lines) in the \textit{low-altitude} phase for wall-effects compensation. From left to right, the blue curves show the $x$-, $y$-, and $\theta$-displacements over time. The red and green curves correspond to the optimized controls $u_1$ and $u_2$, respectively. The yellow plot shows the optimized and classical controls in the phase portrait. Transparency increases with the penalty parameter $\alpha \in \{1, 100, 1000\}$. \textbf{(a)} \textit{Near-field} regime with initial arm length $2R + R/2$. \textbf{(b)} \textit{Middle-field} regime with initial arm length $5R$.}
\label{fig:plotlowaltitude2R5R}
\end{figure*}

\begin{figure*}[htpb]
    \centering
    \includegraphics[width=1\linewidth]{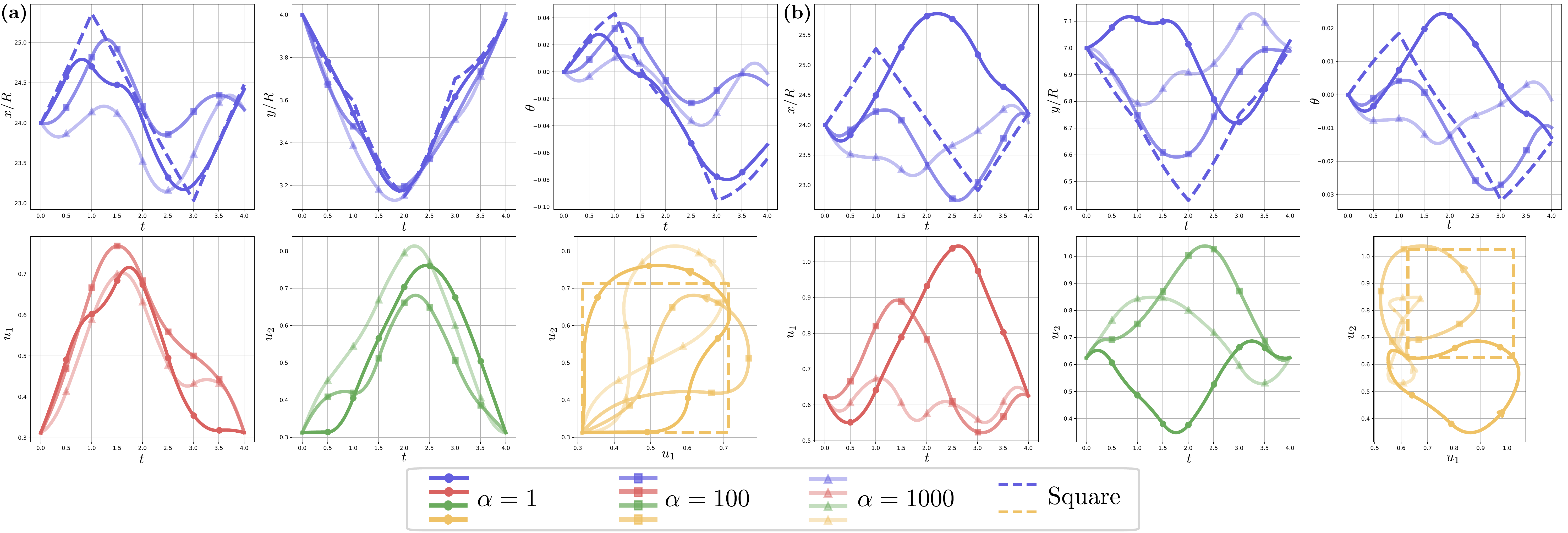}
\caption{Trajectories and control inputs of the optimized swimmers (solid lines) and the classical swimmer (dotted lines) in the \textit{middle-altitude} phase for wall-effects compensation. From left to right, the blue curves show the $x$-, $y$-, and $\theta$-displacements over time. The red and green curves correspond to the optimized controls $u_1$ and $u_2$, respectively. The yellow plot shows the optimized and classical controls in the phase portrait. Transparency increases with the penalty parameter $\alpha \in \{1, 100, 1000\}$. \textbf{(a)} \textit{Near-field} regime with initial arm length $2R + R/2$. \textbf{(b)} \textit{Middle-field} regime with initial arm length $5R$.}
\label{fig:plotmiddlealtitude2R5R}
\end{figure*}

In this section, we employ Bayesian optimization to compensate for the influence of the wall. We focus on the \textit{near-field} and \textit{middle-field} regimes, where the angular deviation is most significant (black stars in \Cref{fig:impact_of_wall_three_sphere}). For each regime, we consider the two opposite behaviours induced by the initial height $h$ above the wall:  
(i) the \textit{low-altitude} phase, where the swimmer tends to move away from the wall, and  (ii) the \textit{middle-altitude} phase, where the swimmer instead drifts toward the wall.  As before, our objective is to maximize the net displacement over one stroke; however, in this setting we additionally seek to compensate the angular drift caused by the wall, as illustrated in \Cref{fig:impact_of_wall_three_sphere}. To this end, we define the following reference final state:
\begin{equation}
    \boldsymbol{p}_{\text{final}}
    = \begin{bmatrix}
        x_{\text{far}} \\
        h \\
        0 \\
        0_{\mathbb{R}^4}
      \end{bmatrix},
\end{equation}
where $x_{\text{far}}$ denotes a far-away target position (here $8R$), and the third component corresponds to the final orientation. Since our goal is to suppress orientation drift, this component is fixed to zero. To explore different drift compensation, we set $Q = 0_{7,7}$ and $Q=\diag([1,\alpha, \alpha, 0, \ldots, 0])$ with $\alpha\in\{1,100, 1000\}$ which penalize deviations in $\theta(T)$ and $y(T)$. The various results obtained are shown in \Cref{fig:plotlowaltitude2R5R}, \Cref{fig:plotmiddlealtitude2R5R} and \Cref{table:costwallcompensation}.\\

\noindent {\bf Low-altitude :} 
In \Cref{fig:plotlowaltitude2R5R}, the results obtained in the \textit{low-altitude} configuration for the first two regimes are presented. 

In the top panel, \textbf{(a)}, corresponding to the \textit{near-field} case, increasing the penalization coefficient $\alpha$ leads to stronger compensation of the angular drift and of the height $y$, to the detriment of net displacement". For $\alpha = 1$ and $\alpha = 100$, the results are similar, which is particularly visible in the phase portraits. For $\alpha = 1000$, the phase portrait becomes much more symmetric and narrow, indicating that the two arms have almost identical deformations throughout the stroke (identical deformations correspond to the line $y = x$ in the phase portrait). This reduces the drift, but decreases the distance traveled. 

The bottom panel, \textbf{(b)}, presents the \textit{middle-field} case. In this regime, the drift is weaker, so the compensation induced by the penalties is less pronounced. In fact, for $\alpha = 1$ and $\alpha = 100$, the drift is even larger than in the classical case, due to a greater net displacement, which compensates its contribution to the objective function. The phase portraits exhibit markedly different shapes depending on the value of $\alpha$.
Using an asymptotic expansion, the dynamics of the three-sphere swimmer can be approximated by an ordinary differential equation that is linear in the control variables \cite{alouges2008, alouges2013, gerardvaret2015}. In this framework, the reachable set is determined by the Lie algebra generated by the iterated Lie brackets of the associated vector fields. For this class of systems, classical control theory tools (see \cite{coron2007}) show that the net displacement over one stroke is proportional to the area enclosed by the gait trajectory in the phase portrait. In the cases studied here, compensatory motions reduce the net displacement, leading to a smaller enclosed area and ensuring that the corresponding gait curves remain contained within the classical stroke pattern.\\

\noindent {\bf Middle-altitude :} In \Cref{fig:plotmiddlealtitude2R5R}, the results obtained in the \textit{middle-altitude} configuration for the first two regimes are presented. 

In the top panel, \textbf{(a)}, corresponding to the \textit{near-field} case, Bayesian optimization identifies controls that compensate the deviations induced by the wall, in particular the angular drift. This compensation becomes more pronounced as the penalty coefficient $\alpha$ increases. However, this causes a reduction in net displacement: the distances achieved for $\alpha = 100$ and $\alpha = 1000$ are nearly identical, but both remain smaller than for $\alpha = 1$. The phase portraits of $\alpha = 100$ and $\alpha = 1000$ highlight nearly identical shapes, differing from a rotation. 

The bottom panel, \textbf{(b)}, presents the \textit{middle-field} case. As already suggested by \Cref{fig:impact_of_wall_three_sphere}, this regime is less sensitive to wall effects, and the resulting compensations also. Nevertheless, the optimization still identifies controls that progressively reduce the deviation as $\alpha$ increases, and their amplitude diminishes with larger penalties. However, despite the very small angular drift obtained for $\alpha = 1000$, the resulting displacement in the $x$-direction becomes extremely small, unlike in \textbf{(a)}, where the swimmer obtained a reasonable progression. The phase portraits also differ across the three values of $\alpha$, in particular with the case $\alpha = 1000$ yielding an eight–shaped loop with two intersections. In both cases, a figure-eight pattern emerges in the periodic trajectory, which reflects the use of a double Lie bracket strategy to maintain the swimming direction. This behavior is consistent with the Lie algebra analysis presented in \cite{alouges2013}.\\

\begin{table}[htpb]
\caption{Cost values for wall effect compensation in the \textit{low-altitude} and \textit{middle-altitude} phases, for both \textit{near-field} and \textit{middle-field} regimes, and for each penalty parameter $\alpha \in \{1,100,1000\}$.}
\label{table:costwallcompensation}
\begin{center}
\begin{tabular}{l l | c c c}
\hline
\hline
Phase & $\begin{array}{l}\text{Regime}\end{array}$ & $\alpha=1$ & $\alpha=100$ & $\alpha=1000$\\
\hline
\textit{Low-altitude} &$\begin{array}{l}
   \textit{Near-field}  \\
     \textit{Middle-field}
\end{array}$ & $\begin{array}{c}0.848\\0.896\end{array}$  & $\begin{array}{c}0.867\\0.921\end{array}$ & $\begin{array}{c}0.953\\0.976\end{array}$\\ 
\hline
\textit{Middle-altitude}  &$\begin{array}{l}\textit{Near-field}  \\\textit{Middle-field} \end{array}$  & $\begin{array}{c}0.902\\0.969\end{array}$  & $\begin{array}{c}0.968\\0.971\end{array}$ & $\begin{array}{c}0.962\\0.992\end{array}$\\ 
\hline
\hline
\end{tabular}
\end{center}
\end{table}

\section{Conclusion and perspectives}\label{sec:conclusion}

The use of high-dimensional Bayesian optimization tools has provided effective results for trajectory tracking across different levels of dynamic complexity. The first model, a flagellated swimmer with a magnetic head, is described by an ODE based on the RFT. The second model, a rigid swimmer composed of three spheres, is described by a PDE solved using finite elements, accounting for interactions with walls. In the first case, the swimmer is able to track trajectories with periodic motion pattern along the path. In the second case, Bayesian optimization produces controls that compensate for angular drift caused by wall hydrodynamical interactions, whether attractive or repulsive.\\

Several extensions are possible. The optimization problem could be made more complex by including the final time or the initial condition as optimizable variables. Transforming the trajectory tracking problem into path tracking is another possibility. In this context, other studies \cite{hung2023} have explored objects capable of following prescribed paths using stabilization methods or Model Predictive Control (MPC). Challenges in this setting include the inability of microswimmers to follow paths exactly, wall or obstacle effects preventing the use of overly simplified models, and the overall system complexity forcing the dynamics to be treated as a black-box function.\\

Our future work aims to increase the physical complexity of the model by incorporating constraints such as swimmer elasticity, the presence of obstacles, or interactions with other swimmers. To address these challenges, Bayesian optimization techniques could be used to efficiently handle this even more challenging setting.


\section*{Acknowledgments}
The authors acknowledge the support of the French Agence Nationale de la Recherche
(ANR) under Grant No.\ ANR-21-CE45-0013 (Project NEMO). 
They also express their gratitude to Christophe Prud'homme for his valuable advice 
regarding the \texttt{Feel++} library.

\appendices

\section{Reminders on rotation matrices}
Let $R \in SO(3)$ be a rotation matrix. We have
\begin{equation}\label{eq:idso3}
    R R^\top = I_3,
\end{equation}
and for all $v \in \mathbb{R}^3$,
\begin{equation}\label{eq:crossmatrix}
    \dot{R} R^\top v = \Omega \times v = [\Omega]^{\times} v,
\end{equation}
where $\Omega \in \mathbb{R}^3$ is the angular velocity associated with $R$, and $[\,\cdot\,]^{\times}$ denotes the matrix representation of the cross product, defined by
\begin{equation*}
    [\Omega]^{\times} =
    \begin{bmatrix}
        0 & -\Omega_3 & \ \Omega_2 \\
        \Omega_3 & 0 & -\Omega_1 \\
        -\Omega_2 & \Omega_1 & 0
    \end{bmatrix},
\end{equation*}
which is antisymmetric.

\section{Matricial form of magnetic swimmer}\label{app:mat_form}
Using the expression of the cross-product matrix introduced in~\eqref{eq:crossmatrix} for the angular velocities $\boldsymbol{\Omega}^h$ and $\boldsymbol{\Omega}^i$, it can be shown, after lengthy but straightforward calculations, that
\begin{equation*}
\boldsymbol{\Omega}^h = L^{h}  \dot{\boldsymbol{\Theta}} \qquad \text{et} \qquad \boldsymbol{\Omega}^i = L^i \begin{bmatrix}
    \dot{\phi}^i_y\\
    \dot{\phi}^i_z
\end{bmatrix},
\end{equation*}
where $L^{h}\in\mathbb{R}^{3\times3}$ et $L^i\in\mathbb{R}^{3\times 2}$ are matrices defined 
{\small
\begin{equation*}
  L^{h}:=\begin{bmatrix}
      1 & 0 & \sin(\theta_y)\\
      0 & \cos(\theta_x) & -\cos(\theta_y)\sin(\theta_x)\\
      0 & \sin(\theta_x) & \cos(\theta_x)\cos(\theta_y)
  \end{bmatrix}   ~\text{and} ~L^i:=\begin{bmatrix}
      0 & \sin(\phi^i_y)\\
      1 & 0\\
      0 &\cos(\phi_y^i)
  \end{bmatrix}.
\end{equation*}
}
\medskip

\noindent We now introduce the coefficients of a block matrix $A \in \mathbb{R}^{3m \times 3m}$ such that each block
$A_{[i,j]} = A_{3(i-1)+1:3i,\, 3(j-1)+1:3j}$ 
represents the $(i,j)$-th submatrix of size $(3\times3)$.  
For a matrix $B \in \mathbb{R}^{3m\times3}$, we denote
$B_{[i,]} = B_{3(i-1)+1:3i,\, 1:3}$, 
and for a matrix $C \in \mathbb{R}^{3\times3m}$, we write
$C_{[,j]} = C_{1:3,\, 3(j-1)+1:3j}$. Using~\eqref{eq:Oi} and its time derivative, we define the matrices
$B \in \mathbb{R}^{(3N+6)\times(N+6)}$ and
$Q \in \mathbb{R}^{(6N+6)\times(3N+6)}$ as follows:
\begin{equation*}
B=
\left[
\begin{array}{cc|ccc}
I_3 & 0 & 0 & \cdots & 0 \\
0 & L^{h} & 0 & \cdots & 0 \\\hline
0 & 0 & L^1 & \cdots & 0 \\
\vdots & \vdots & \vdots & \ddots & \vdots \\
0 & 0 & 0 & \cdots & L^N
\end{array}
\right]
\end{equation*}
and 
\begin{equation*}
\quad Q =
\left[
\begin{array}{c|c|c}
I_3 & 0_3 & \begin{matrix}0_3 & \cdots & 0_3 \end{matrix}\\ \hline
\begin{matrix} I_3 \\ \vdots \\ I_3\end{matrix} &   q^{h} & q^L  \\\hline
0_3 & I_3 & \begin{matrix}0_3 & \cdots & 0_3\end{matrix}\\\hline
\begin{matrix} 0_3\\ \vdots \\ 0_3 \end{matrix} & \begin{matrix} 0_3 \\\vdots \\0_3\end{matrix}  &  I_{3N} \\
\end{array}
\right],
\end{equation*}
where, for $1 \leq i \leq N$ and $1 \leq j \leq i-1$,
\begin{equation*}
\begin{cases}
\left(q^{h}\right)_{[i,]} = r [R^h \boldsymbol{e}_1]^{\times} 
+ l \displaystyle\sum_{k=1}^{i-1} [R^h R^k \boldsymbol{e}_1]^{\times}, \\
\left(q^{L}\right)_{[i,j]} = l R^h [R^j \boldsymbol{e}_1]^{\times}.
\end{cases}
\end{equation*}
Finally, the matrices $Q$ and $B$ are related through :
\begin{equation*}
    QB  
    \begin{bmatrix}
        \dot{\boldsymbol{X}}\\
        \dot{\boldsymbol{\Theta}}\\
        \dot{\boldsymbol{\Phi}}
    \end{bmatrix} = 
    \begin{bmatrix}
        \dot{\boldsymbol{X}}\\
        \dot{\boldsymbol{X^1}}\\
        \vdots\\
        \dot{\boldsymbol{X^N}}\\
        \boldsymbol{\Omega}^h\\
        \boldsymbol{\Omega^1}\\
        \vdots\\
        \boldsymbol{\Omega^N}
    \end{bmatrix}.
\end{equation*}

\medskip

\noindent Using the expressions of the hydrodynamic forces and torques, we can define the global system matrix $A$ of dimension $(2N+6)\times(6N+6)$ as
\begin{equation*}
A = 
\begin{bmatrix}
A^{F,\dot{X}_{\text{head}}} & A^{F,\dot{X}_{\text{links}}} & A^{F,\Omega^h} & A^{F,\Omega} \\
0_{3,3} & A^{T_{\text{total}},\dot{X}_{\text{links}}} & A^{T_{\text{total}},\Omega^h} & A^{T_{\text{total}},\Omega} \\
0_{2N,3} & A^{T_{\text{sub}},\dot{X}_{\text{links}}} & 0_{N,3} & A^{T_{\text{sub}},\Omega}
\end{bmatrix},
\end{equation*}
where the individual block matrices are given by
{\small
\begin{equation*}
\begin{cases}
A^{F,\dot{X}_{\text{head}}} = - R^h D^h (R^h)^\top r,\\
A^{F,\Omega^h} = -\dfrac{l^2}{2}\displaystyle\sum_{i=1}^N R^h \tilde{D}^i (R^h)^\top [R^h R^i \boldsymbol{e}_1]^{\times},\\
A^{T_{\text{total}},\Omega^h} =
\begin{multlined}[t]
 -k_r r^3 I_3 + l^2 \displaystyle\sum_{i=1}^N 
 \left(\dfrac{l}{3}[R^h R^i \boldsymbol{e}_1]^{\times} 
 - \dfrac{1}{2}[\boldsymbol{X}^i - \boldsymbol{X}]^{\times}\right) \\
 R^h \tilde{D}^i (R^h)^\top [R^h R^i \boldsymbol{e}_1]^{\times}.
\end{multlined}
\end{cases}
\end{equation*}
}

For $j = 1,\ldots,N$, we define the following block matrices, each constructed by concatenating submatrices columnwise:
{\small
\begin{equation*}
\begin{cases}
\left(A^{F,\dot{X}_{\text{links}}}\right)_{[,j]} = - R^h \tilde{D}^j (R^h)^\top l,\\
\left(A^{F,\Omega}\right)_{[,j]} = - R^h \tilde{D}^j [R^j \boldsymbol{e}_1]^{\times} \dfrac{l^2}{2},\\
\left(A^{T_{\text{total}},\dot{X}_{\text{links}}}\right)_{[,j]} =
\begin{multlined}[t]
 l\left(\dfrac{l}{2}[R^h R^j \boldsymbol{e}_1]^{\times} 
 - [\boldsymbol{X}^j - \boldsymbol{X}]^{\times}\right)
 R^h \tilde{D}^j (R^h)^\top,
\end{multlined}\\
\left(A^{T_{\text{total}},\Omega}\right)_{[,j]} =
\begin{multlined}[t]
 l^2 \left(\dfrac{l}{3}[R^h R^j \boldsymbol{e}_1]^{\times} 
 - \dfrac{1}{2}[\boldsymbol{X}^j - \boldsymbol{X}]^{\times}\right)
 R^h \tilde{D}^j [R^j \boldsymbol{e}_1]^{\times}.
\end{multlined}
\end{cases}
\end{equation*}
}
For each $i \in \{1, \ldots, N\}$ and for all $j$ such that $i \leq j \leq N$, we define the following upper triangular block matrices:
{\small
\begin{equation*}
\begin{cases}
(A^{T_{\text{sub}},\dot{X}_{\text{links}}})_{[i,j]} =
\begin{multlined}[t]
 l\, \Pi^i
 \left(\dfrac{l}{2}[R^h R^j \boldsymbol{e}_1]^{\times}
 - [\boldsymbol{X}^j - \boldsymbol{X}^i]^{\times}\right)
 R^h \tilde{D}^j (R^h)^\top,
\end{multlined}\\
(A^{T_{\text{sub}},\Omega})_{[i,j]} =
\begin{multlined}[t]
 l^2 \Pi^i 
 \left(\dfrac{l}{3}[R^h R^j \boldsymbol{e}_1]^{\times}
 - \dfrac{1}{2}[\boldsymbol{X}^j - \boldsymbol{X}^i]^{\times}\right)
 R^h \tilde{D}^j [R^j \boldsymbol{e}_1]^{\times}.
\end{multlined}
\end{cases}
\end{equation*}
}
with $\boldsymbol{X}^j$ defined by~\eqref{eq:Oi}.

\medskip

\noindent The constant elastic contribution $\boldsymbol{F}_0$ is defined as
\begin{equation*}
\boldsymbol{F}_0 
= k_{\text{el}}
\begin{bmatrix}
0_6\\
\Pi^1 [R^h R^1 \boldsymbol{e}_1]^{\times} R^h \boldsymbol{e}_1\\
\vdots\\
\Pi^i [R^h R^i \boldsymbol{e}_1]^{\times} R^h R^{i-1} \boldsymbol{e}_1\\
\vdots
\end{bmatrix},
\end{equation*}
and for each control input $i$, the magnetic torque term $\boldsymbol{F}_i$ is expressed as
\begin{equation*}
\boldsymbol{F}_i 
= m
\begin{bmatrix}
0_3\\
[R^h \boldsymbol{e}_1]^{\times} \boldsymbol{e}_i\\
0_{2N}
\end{bmatrix}.
\end{equation*}

\section{Comparison between 2D and 3D three-sphere models}\label{app:threesphere2Dvs3D}

\begin{figure}[htpb]
    \centering
    \includegraphics[width=0.7\linewidth]{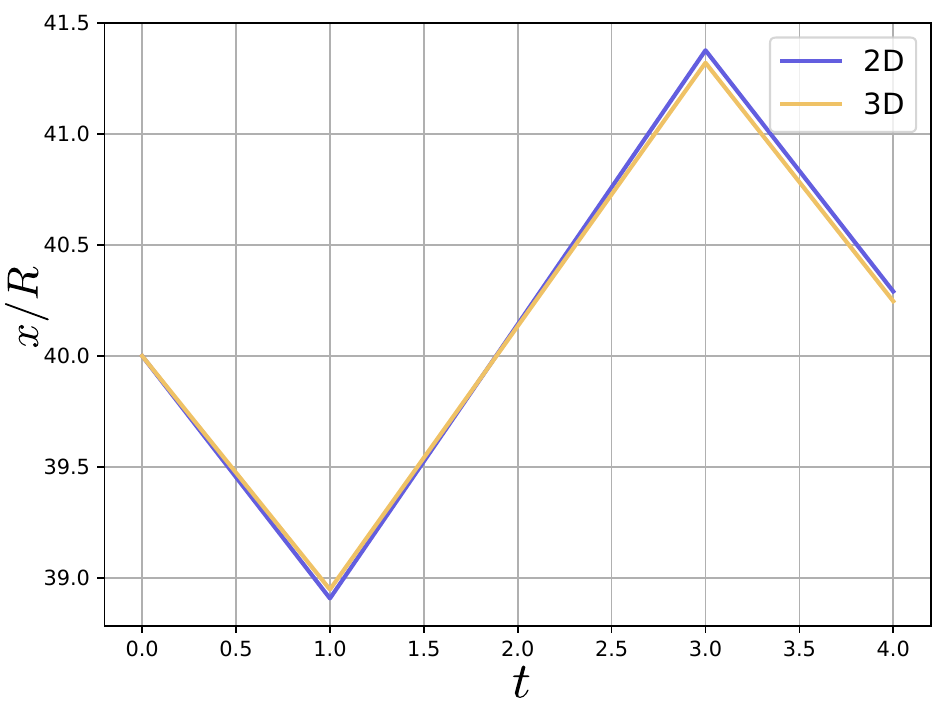}
    \caption{Net $x$–displacement of the three-sphere swimmer in 2D (blue) and 3D (yellow). Initial arm length equal to $10R$.}
    \label{fig:threesphere2Dvs3D}
\end{figure}

For the same radius $R$, an initial arm length of $10R$, and identical actuation of the two arms, \Cref{fig:threesphere2Dvs3D} compares the net $x$–displacement of the 2D and 3D three-sphere swimmers. At the end of the stroke, the absolute difference between the two displacements is $0.15R$.

\section{Impact of the number of control points}\label{app:cost_vs_nbctrl}

\begin{figure}[htpb]
    \centering
    \includegraphics[width=0.7\linewidth]{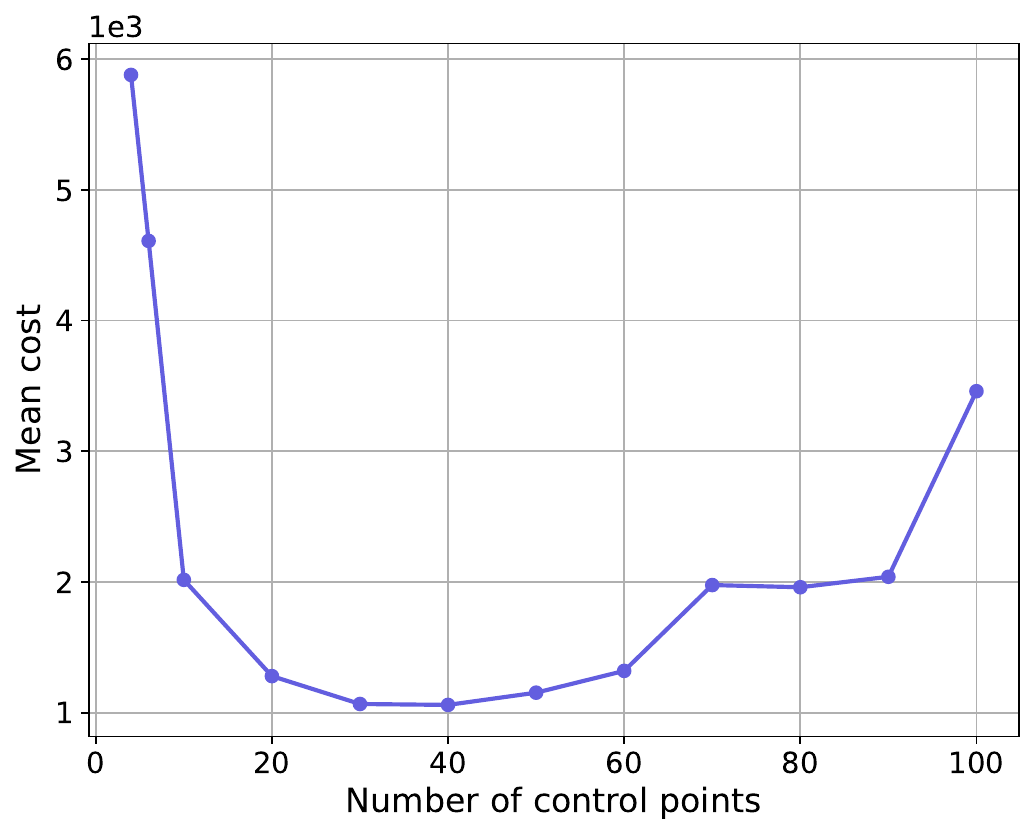}
   \caption{Mean optimal cost obtained for the elliptical trajectory tracking of the $N$-link swimmer with $a = L$ and $b = L/2$, as a function of the number of control points. The mean is computed over three independent optimization runs.}
    \label{fig:cost_vs_nbctrl}
\end{figure}

To analyze the influence of the number of control points on the search for optimal controls in trajectory tracking, we analyze the optimal cost obtained after optimization as a function of the number of control points, the problem dimension. We focus on the $N$-link swimmer tracking an elliptical trajectory of total length $L$, with radii $a = L$ and $b = L/2$, as defined in \eqref{eq:ellipse_traj} and illustrated in \Cref{fig:N_link_2D_ellipses} \textbf{(a)}. For each Bayesian optimization run, the number of cost evaluations is fixed to $2000$. The resulting optimal costs for different numbers of control points are reported in \Cref{fig:cost_vs_nbctrl}.

\section{Optimization parameters}
The SCBO parameters for all optimization problems are listed in \Cref{table:param_scbo}.
\begin{table}[htpb]
\caption{SCBO parameters.}
\label{table:param_scbo} 
\begin{center}
  \begin{tabular}{l c l}
  \hline
\hline
     Parameter & Value & Description  \\ 
     \hline
     N & $-$ & Problem dimension.\\
     $r$  & $\min(5000, \max(2000,200\times N))$ & Candidates in region \\ 
     $q$ & $-$ & Batch size \\ 
     $L_{\text{init}}$ & $1.6$ & Initial region length \\ 
     $L_{\text{min}}$ & $0.5^{7}$ & Min region length \\ 
     $L_{\text{max}}$ & $1.6$ & Max region length \\ 
     $n_{\text{init}}$ & $4\times N$ & Initial points \\  
    $\tau_s$ & $\max(3, \lceil N/10\rceil)$& Success rate.\\
    $\tau_f$ & $\lceil N/q \rceil$ & Failure rate.\\
    \hline
\hline
  \end{tabular}
  \end{center}
\end{table}

\bibliographystyle{IEEEtran}
\bibliography{biblio}

\end{document}